\documentclass[iicol,pdflatex,sn-mathphys-ay]{sn-jnl}% Math and Physical Sciences Author Year Reference Style
\usepackage{graphicx}%
\usepackage{multirow}%
\usepackage{amsmath,amssymb,amsfonts}%
\usepackage{amsthm}%
\usepackage{mathrsfs}%
\usepackage[title]{appendix}%
\usepackage{xcolor}%
\usepackage{textcomp}%
\usepackage{manyfoot}%
\usepackage{booktabs}%
\usepackage{threeparttable}%
\usepackage{algorithm}%
\usepackage{algorithmicx}%
\usepackage{algpseudocode}%
\usepackage{listings}%
\hypersetup{hypertexnames=false,bookmarksdepth=2}
\theoremstyle{thmstyleone}%
\theoremstyle{thmstyletwo}%

\theoremstyle{thmstylethree}%

\begin{document}

\title[CoVar for Pseudo-Label Selection]{CoVar: Confidence--Variance-Guided Pseudo-Label Selection for Semi-Supervised Learning}


\author[1]{\fnm{Jinshi} \sur{Liu}}

\author*[2]{\fnm{Lei} \sur{He}}

\author*[3]{\fnm{Pan} \sur{Liu}}

\email{helei\_xb@hnust.edu.cn}
\email{liup28292@gmail.com}

\affil[1]{\orgdiv{College of Artificial Intelligence}, \orgname{Shenzhen University}, \orgaddress{\city{Shenzhen}, \postcode{518060}, \country{China}}}

\affil*[2]{\orgdiv{School of Information and Electrical Engineering}, \orgname{Hunan University of Science and Technology}, \orgaddress{\city{Xiangtan}, \postcode{411100}, \country{China}}}

\affil*[3]{\orgdiv{Information Hub}, \orgname{Hong Kong University of Science and Technology (Guangzhou)}, \orgaddress{\city{Guangzhou}, \postcode{510006}, \country{China}}}


\abstract{
Pseudo-label selection in semi-supervised learning is commonly driven by maximum-confidence thresholds, yet confidence alone can be unreliable under model overconfidence and class imbalance.
We propose CoVar, a confidence--variance framework that assesses pseudo-label reliability by jointly modeling Maximum Confidence (MC) and Residual-Class Variance (RCV).
Starting from entropy minimization, we derive a second-order cross-entropy approximation showing that low-loss pseudo-labels are favored when MC is high and RCV is low, with a confidence-dependent penalty that becomes stronger for near-certain predictions.
Based on this criterion, CoVar embeds predictions into a two-dimensional confidence--variance space and uses SVD-based spectral relaxation to separate reliable and unreliable predictions without hand-tuned confidence thresholds.
Cluster-wise Gaussian weighting then converts this separation into per-sample training weights.
The resulting weights can be integrated into existing semi-supervised segmentation and classification pipelines during training and introduce no inference-time overhead.
Experiments on PASCAL VOC 2012, Cityscapes, CIFAR-10, CIFAR-100, SVHN, and STL-10 show clear gains on VOC and Cityscapes under matched backbones, as well as competitive or improved error rates on standard classification benchmarks.
These results indicate that residual-class dispersion provides a useful signal complementary to confidence for robust pseudo-label selection.
}




\maketitle

\section{Introduction}\label{sec1}

\begin{figure}[t]
  \centering
  \includegraphics[width=\linewidth]{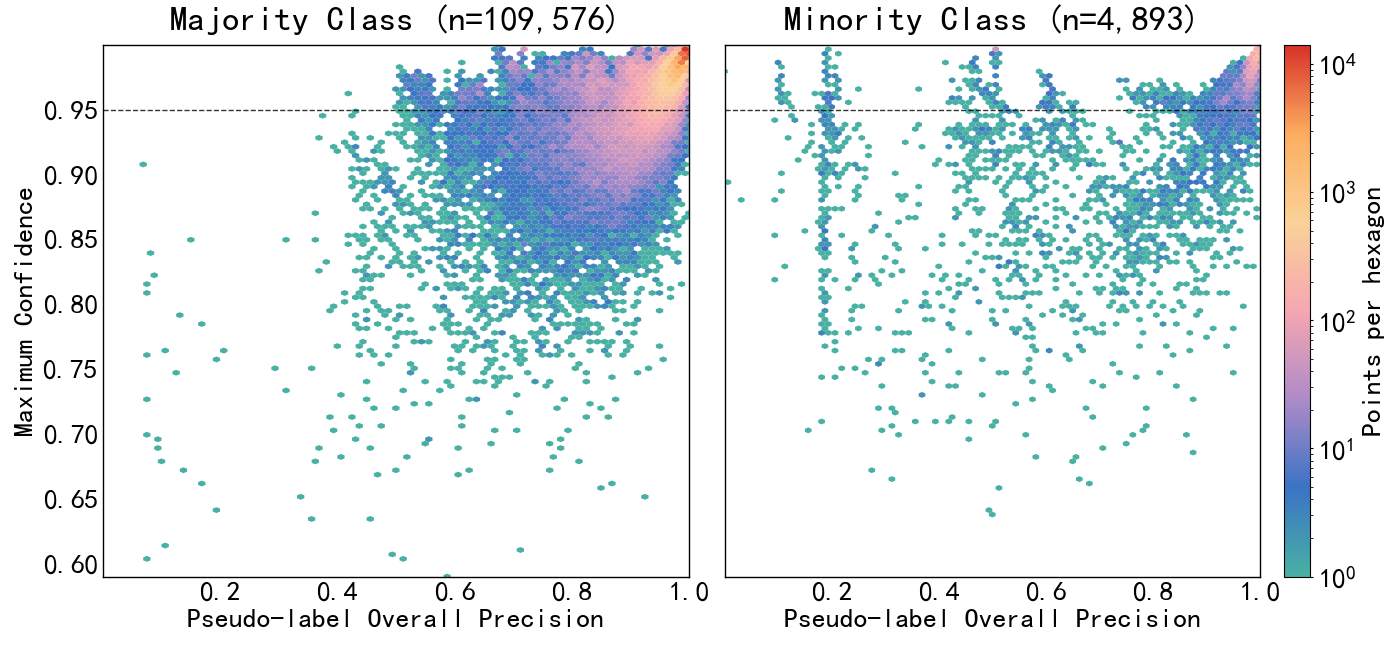}
  \caption{Relationship between pseudo-label precision and maximum confidence on PASCAL VOC 2012 (1/4 split).
  The left and right panels show majority-class and minority-class predictions, respectively. Each hexagonal bin aggregates pixel samples collected during training from epoch 1 to 79, and the color indicates the number of pixels in that bin.
  The horizontal axis reports pseudo-label precision measured on a labeled diagnostic subset used only for offline analysis; it does not participate in parameter updates or model selection. The vertical axis reports maximum confidence.
  Traditional fixed-threshold methods are constrained by the limited separability of confidence scores, which can cause some high-quality predictions to be discarded while some erroneous predictions are retained, ultimately limiting pseudo-label precision.}  
  \label{fig1}
\end{figure}

Semi-supervised learning (SSL) uses a small labeled set together with abundant unlabeled data to reduce dependence on manual annotation.
Pseudo-labeling, a central paradigm in SSL, derives supervisory signals from the predictions of the model and has been widely adopted in image classification and semantic segmentation~\citep{fixmatch,UniMatch,Unimatchv2}.
However, most existing methods select pseudo-labels by applying high-confidence thresholds, implicitly assuming a strong correlation between prediction confidence and correctness~\citep{fixmatch,flexmatch,freematch,softmatch,UniMatch}.
This assumption lacks firm theoretical support.
Manually specified confidence thresholds can introduce additional pseudo-label noise.
Such thresholds can also amplify the overconfidence of deep models trained with limited supervision.
Consequently, predictions often concentrate in extreme confidence ranges while remaining weakly correlated with actual accuracy (Fig.~\ref{fig1}).
This overconfidence has direct practical drawbacks.
Miscalibration allows erroneous predictions to appear as confident as correct ones, weakening confidence as a selection signal and making thresholded pseudo-labels noisy.
Once selected, these high-confidence errors are reinforced by the unsupervised objective and can propagate confirmation bias during training.

Existing threshold-based methods therefore evaluate pseudo-label reliability with a confidence-only criterion.
As illustrated in Fig.~\ref{fig1}, high-confidence pseudo-labels are not necessarily high-precision pseudo-labels, especially when confidence scores have limited separability.
This observation motivates a principled reliability criterion defined directly on model predictions.
Such a criterion should separate trustworthy pseudo-labels from unreliable ones in a systematic way.
To this end, we propose a unified Confidence--Variance (CoVar) theoretical framework that generalizes confidence-threshold filtering by incorporating residual-class dispersion.
The core theoretical contribution is a reliability criterion that mitigates overconfidence and improves pseudo-label selection.
Starting from entropy minimization, our analysis shows that a reliable pseudo-label should satisfy two coupled requirements: sufficiently high Maximum Confidence (MC) and low Residual-Class Variance (RCV).
In multi-class settings, RCV measures the dispersion of the residual-class distribution, and low RCV indicates that no non-maximum class forms a strong competing hypothesis.
The derivation further reveals a dynamic interaction between the two factors: the RCV penalty increases with maximum confidence.
This confidence-dependent penalty requires a high-confidence prediction to have correspondingly low RCV before it is considered reliable.
Under the resulting second-order analysis, low-loss predictions are therefore characterized by high MC and low RCV jointly.

The main contributions are threefold.

\begin{enumerate}
\item We derive a second-order CE reliability criterion with a practical adaptive residual target, showing that pseudo-label reliability depends jointly on MC and RCV and yielding a batch-level explanation for long-tail bias mitigation.
\item We propose CoVar, a training-time SVD spectral selection module that separates predictions in confidence--variance space and converts the separation into Gaussian sample weights without hand-tuned thresholds or inference-time overhead.
\item We validate CoVar on semi-supervised segmentation and classification, with ablations confirming the roles of RCV, confidence-dependent weighting, and spectral separation.
\end{enumerate}

This article extends our ICCV paper CSL~\citep{Liu_2025_ICCV}, which introduced the core confidence-based pseudo-label selection framework.
The present work further develops the second-order analysis of pseudo-label reliability, extends the formulation to a batch-level view of class-imbalance effects, refines the spectral selection strategy, and broadens the empirical study with image classification benchmarks, additional ablations, runtime analysis, and a unified presentation of the CoVar framework.

\section{Related Work}\label{sec2}

Semi-supervised learning~\citep{Unsupervised,kim2022conmatch,Usb,Liu_2025_ICCV} uses pseudo-labels to transfer supervision from labeled to unlabeled data, but pseudo-label quality is often limited by confirmation bias, overconfidence, and class imbalance~\citep{xie2020self,zoph2020rethinking,ljs2,MT,fixmatch}. Existing work can be grouped into five closely related directions.

\subsection{Confidence-Threshold Methods}

The dominant family of SSL methods selects pseudo-labels using maximum confidence, either with fixed thresholds or adaptive threshold schedules. Representative examples include FixMatch~\citep{fixmatch}, FlexMatch~\citep{flexmatch}, FreeMatch~\citep{freematch}, and SoftMatch~\citep{softmatch} for image classification, as well as UniMatch and UniMatch~V2 for semantic segmentation~\citep{UniMatch,Unimatchv2}. These methods improve the quantity--quality tradeoff of pseudo-labels, but they still treat confidence as the primary reliability signal. As a result, predictions with identical maximum confidence but different residual-class structures are often indistinguishable to threshold-based rules, whereas CoVar explicitly models residual-class dispersion to separate such cases.

\subsection{Calibration-Based Pseudo-Labeling}

Another line of work improves pseudo-label selection by calibrating predictive confidence. Classical post-hoc calibration methods include temperature scaling~\citep{guo2017calibration}, Platt scaling~\citep{platt1999probabilistic}, and histogram binning~\citep{zadrozny2001obtaining}. Training-time techniques such as label smoothing~\citep{szegedy2016rethinking}, focal loss~\citep{lin2017focal}, Mixup~\citep{zhang2017mixup}, and CutMix~\citep{yun2019cutmix} also alleviate overconfidence, while related settings such as domain adaptation adopt transferable calibration strategies~\citep{PC,MIC}. These methods improve the alignment between confidence and accuracy, but they still summarize reliability mainly through scalar confidence or entropy and do not explicitly model the distribution of residual non-maximum probabilities. In particular, they do not distinguish concentration of residual mass from total uncertainty, whereas CoVar makes this distinction explicit through RCV.

\subsection{Uncertainty- and Variance-Aware Selection}

Beyond confidence calibration, several methods incorporate richer reliability signals such as uncertainty, weighting, or distributional dispersion. Examples include Monte Carlo dropout~\citep{defense}, ensemble disagreement~\citep{Double}, reward-based filtering~\citep{SemiReward}, pixel-wise weighting strategies such as AEL and DAW~\citep{AEL,DAW}, multi-hot pseudo-labeling in ESL~\citep{ESL}, and contrastive treatment of unreliable pixels in U2PL~\citep{U2PL}. These approaches show that confidence alone is insufficient, but they typically rely on auxiliary models, additional heuristics, or task-specific weighting rules. CoVar differs by deriving a residual-class variance term directly from the entropy-minimization objective and coupling it analytically with confidence, rather than introducing another auxiliary uncertainty estimator or handcrafted reliability rule.

\subsection{Spectral Clustering and Graph Partition in SSL}

Graph-based and partition-based ideas have also been used to improve SSL. In segmentation, CorrMatch~\citep{CorrMatch} propagates labels through correlation matching, showing that relationships among predictions can be more informative than independent thresholding. More broadly, normalized partition and spectral relaxation provide principled tools for separating data according to global structure~\citep{relaxation,kf}. CoVar is related to this line only at the level of relaxation: it does not cluster pixels, images, or feature graphs, but instead performs a two-way separation on prediction-reliability statistics in a two-dimensional confidence--variance space derived from the CE decomposition.

\subsection{Class Imbalance and Long-Tail Pseudo-Label Bias}

Recent work has shown that pseudo-label selection is often biased toward head classes or easy majority patterns. DARS explicitly redistributes biased pseudo-labels in semi-supervised segmentation~\citep{DARS}, and debiased self-training addresses similar issues in general SSL~\citep{chen2022debiased}. Related segmentation methods also report that confidence-based rules can under-select reliable minority-class pixels even when the corresponding predictions are correct~\citep{CorrMatch,U2PL,DAW}. CoVar addresses this issue from a different perspective: it is not an explicit class-rebalancing method and does not introduce class-frequency reweighting or pseudo-label redistribution, but instead uses batch-level coupling between confidence and residual-class variance to explain and mitigate long-tail selection bias.

In summary, prior work has improved pseudo-label selection through thresholding, calibration, uncertainty estimation, graph propagation, and debiasing. CoVar is closest in spirit to uncertainty-aware and partition-based selection, but differs in three ways: it derives reliability from a second-order CE decomposition rather than heuristic confidence surrogates, it separates prediction-reliability statistics in a confidence--variance space via spectral relaxation rather than clustering image or feature graphs or relying on confidence thresholding alone, and it connects this mechanism to batch-level bias mitigation under class imbalance without explicit class reweighting.

\section{Methodology}

\subsection{Preliminaries}

\paragraph{Pseudo-Label Selection in Semi-Supervised Learning}

In semi-supervised learning, the training data consists of a labeled set $D_l={\{({x^l_i},{y^l_i})\}}^{N_l}_{i=1}$ and an unlabeled set $D_u={\{({x^u_j})\}}^{N_u}_{j=1}$, where the number of labeled samples \(N_l\) is much smaller than the number of unlabeled samples \(N_u\) (i.e., \(N_l \ll N_u\)).
The goal of SSL is to learn a model $ f_{\theta}:x\rightarrow y$ by minimizing a weighted combination of supervised and unsupervised losses:
\begin{equation}
  {L}_{\mathrm{total}}={{L}_{l}}({{D}_{l}})+\lambda {{L}_{u}}({{D}_{u}})    
    \label{eq2}
\end{equation}
where \(L_u(D_u)\) denotes the unsupervised loss computed on the unlabeled dataset \(D_u\).
This loss uses pseudo-labels \(\hat{y}_{j}^{u}\) generated for unlabeled samples \(x_{j}^{u}\) as supervisory signals.
The coefficient \(\lambda\) is a balancing hyperparameter.
Meanwhile, \(L_l(D_l)\) denotes the supervised loss on the labeled dataset \(D_l\), which guides the model to learn discriminative information from labeled examples.
\begin{equation}
    {{L}_{l}}({{D}_{l}})=\frac{1}{{{N}_{l}}}\sum\limits_{j=1}^{{{N}_{l}}}{\mathcal{L}(y_{j}^{l},{{f}_{\theta }}(x_{j}^{l}))}
    \label{eq3}
\end{equation}
where ${\mathcal{L}}(\cdot, \cdot)$ denotes the loss function, e.g., cross-entropy (CE).
Current confidence-based pseudo-label selection methods, such as FixMatch~\citep{fixmatch}, primarily filter pseudo-labels with a confidence threshold.
Specifically, for each unlabeled sample $x_{j}^{u}$, the model outputs a predicted class probability distribution \( {{p}_{j}}={{f}_{\theta }}(x_{j}^{u}) \), also referred to as a confidence distribution.
If the maximum probability in this distribution exceeds a predefined confidence threshold \( \tau \), the corresponding class is selected as the pseudo-label $\hat{y}_{j}^{u}$; otherwise, the sample is excluded from the unsupervised loss.
\begin{equation}
\hat{y}_{j}^{u}=\left\{ \begin{matrix}
   {{k}'} & \text{if } \max_{k} {{p}_{j}}(k)\ge \tau   \\
   \text{ignore} & \text{otherwise}  \\
\end{matrix} \right.
    \label{eq4}
\end{equation}
where \( {p}_j = \{p_{j}(0), p_{j}(1), \ldots ,p_{j}(K-1)\}\) denotes the confidence distribution, $K$ denotes the number of classes, and \( {k}'=\arg\max_{k} \,{{p}_{j}}(k) \) denotes the class with the highest confidence.
\( \tau \in [0,1] \) is a predefined fixed threshold whose value may vary across tasks.
The unsupervised loss \( L_u(D_u) \) is then constructed by computing the CE between the model prediction \( {p}_j \) and the selected pseudo-label $\hat{y}_{j}^{u}$ under a loss mask.
This mask ensures that only samples satisfying the confidence-threshold condition contribute to the unsupervised loss:
\begin{equation}
  {{L}_{u}}({{D}_{u}})=\frac{1}{{{N}_{u}}}\sum\limits_{j=1}^{{{N}_{u}}}{I(\max_{k} {{p}_{j}}(k)\ge \tau )}\cdot \mathcal{L}(\hat{y}_{j}^{u},{{f}_{\theta }}(x_{j}^{u}))    
    \label{eq5}
\end{equation}
where \( {I}(\cdot) \) is an indicator function.
\begin{figure}[t]
  \centering
    \includegraphics[width=\linewidth]{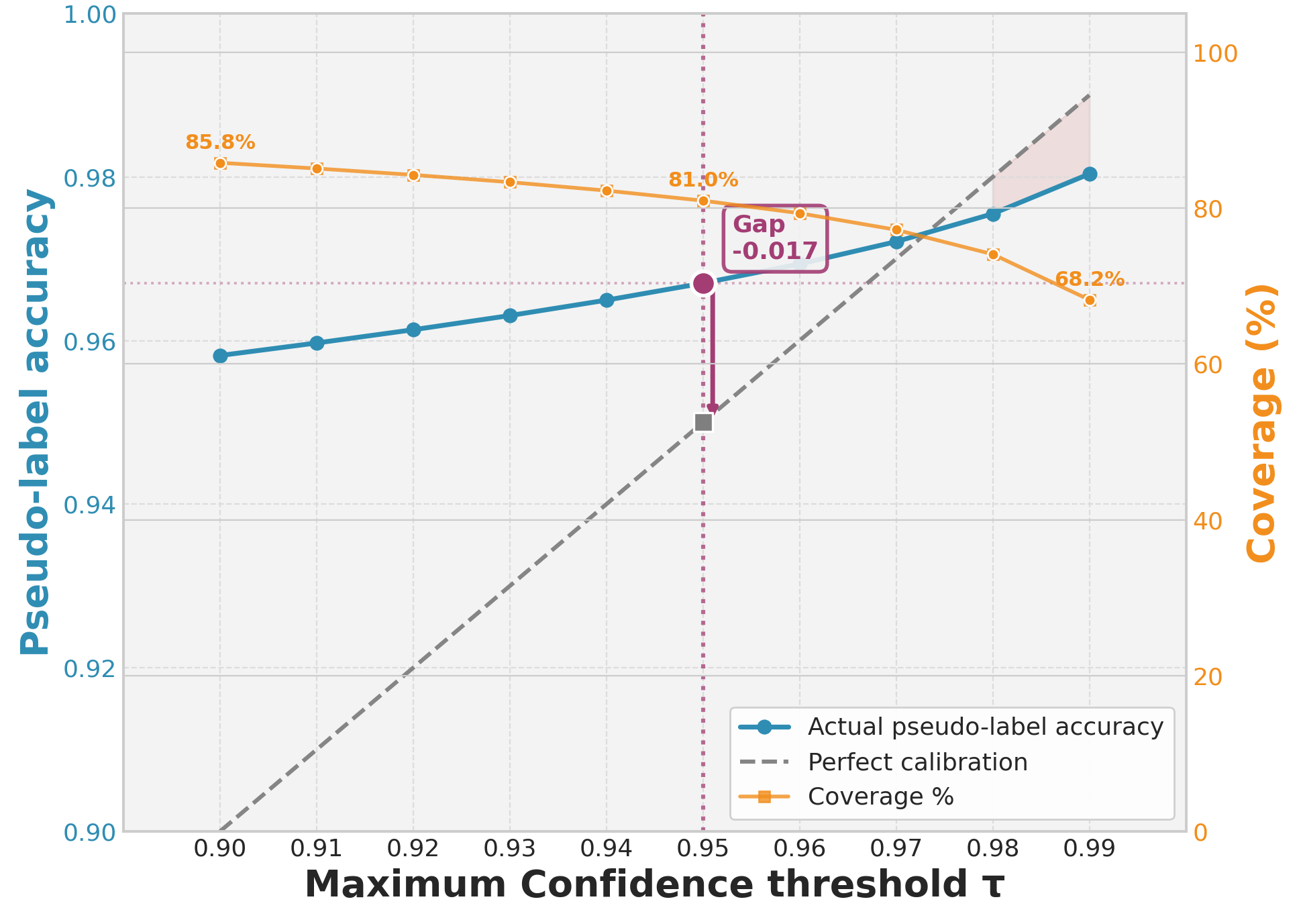}
    \caption{Pseudo-label accuracy and selection ratio under different confidence thresholds on PASCAL VOC 1/4.
    The blue curve shows pseudo-label accuracy under a fixed-threshold selection strategy, the orange curve shows the selected-sample ratio, and the gray dashed line indicates ideal calibration.}
    \label{fig2}
\end{figure}
\paragraph{Limitation of Fixed Threshold: An Accuracy--Coverage Tradeoff}

As illustrated in Fig.~\ref{fig2}, fixed-threshold selection inherently suffers from an accuracy--coverage tradeoff.
When the confidence threshold $\tau$ is relatively low, many pseudo-labels are selected, 
but their accuracy is limited by the inclusion of unreliable predictions. 
Conversely, when $\tau$ is increased to enforce higher precision, pseudo-label accuracy improves, 
but the number of selected samples drops sharply, resulting in insufficient supervisory signals.

This tradeoff indicates that no single fixed threshold can simultaneously achieve high pseudo-label accuracy and high sample coverage~\citep{guo2017calibration}.
In practice, the model is forced into a suboptimal regime:
it either learns from noisy pseudo-labels or receives too little effective training data. 
Therefore, the performance of fixed-threshold methods is highly sensitive to the choice of $\tau$ 
and fundamentally constrained by the accuracy--coverage dilemma.

The gap between the blue curve and the ideal calibration line at a typical threshold (e.g., $\tau_0=0.95$) visualizes the calibration error $\Delta = \lvert \mathbb{P}(\hat{y}=y \mid \max(p)\ge\tau_0)-\tau_0 \rvert$, showing that high-confidence predictions remain systematically overconfident under fixed-threshold selection.

\subsection{Confidence--Variance Theoretical Framework}

\paragraph{Pseudo-Label Selection at the Sample Level}
\begin{figure*}[t]
  \centering
    \includegraphics[width=\linewidth]{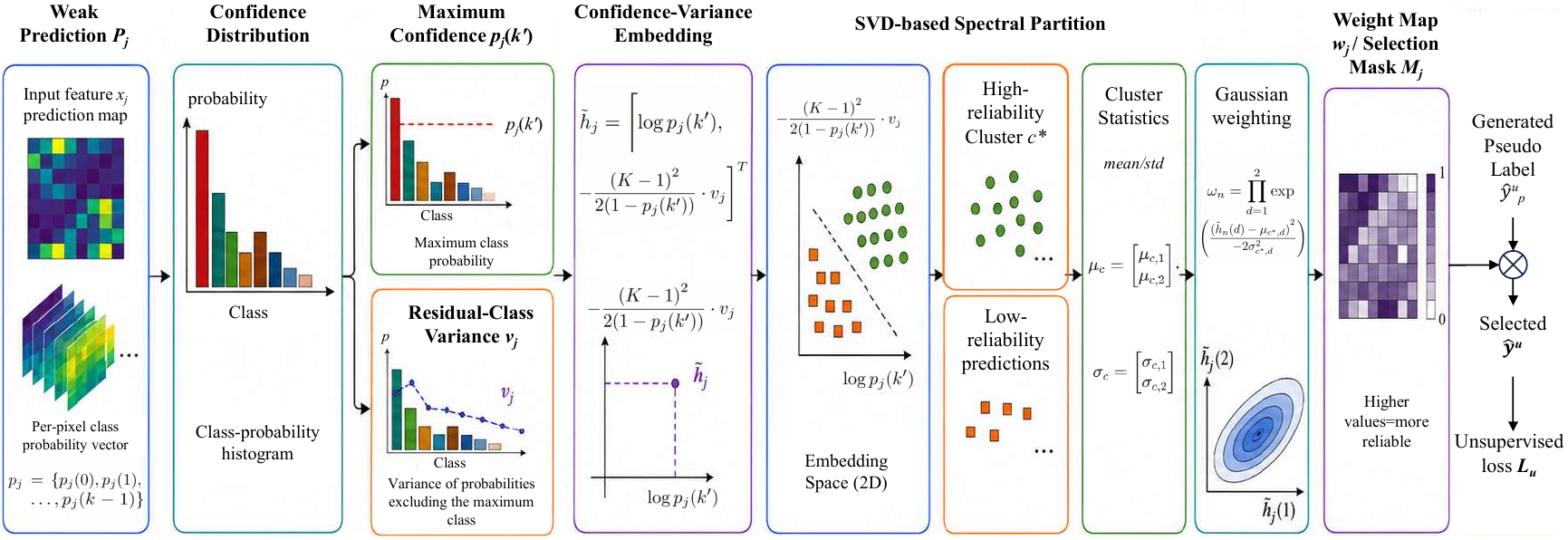}
    \caption{The overall pipeline of the proposed method.
    Based on the proposed confidence--variance theory, the prediction separation module divides pseudo-labels into high-reliability and low-reliability groups.
    Then, Gaussian weighting is applied to transform the statistical characteristics of MC and RCV into pseudo-label weights, enabling the selection of high-quality pseudo-labels.}
    \label{fig3}
\end{figure*}
Entropy minimization is a core strategy in semi-supervised learning and improves model training by reducing prediction uncertainty on unlabeled data.
The central idea is to encourage high-confidence predictions for unlabeled samples, thereby pushing the decision boundary away from high-density data regions.
After prediction on unlabeled data, high-confidence, low-entropy samples are selected to generate pseudo-labels and are then used for iterative optimization.
Entropy minimization is typically incorporated as an additional unsupervised term in the training objective and optimized jointly with the supervised loss, acting as a regularizer that encourages low-entropy predictions on unlabeled data.

Consider a batch containing \( N_B \) unlabeled samples and their predictions, where \( {p}_j \) denotes the confidence (probability) distribution for the \( j \)-th sample.
Let \( {{p}_{j}}({k}') \) be the Maximum Confidence (MC), and let \( {k}'=\arg\max_{k} \,{{p}_{j}}(k) \) be the class corresponding to this maximum value.
Under entropy minimization, with \(k'\) taken from the current prediction \(p_j\), an ideal target distribution for sample \(j\) can be represented as 
\begin{equation}
q_j(k)=\left\{ \begin{matrix}
   1-(K-1)\varepsilon  & \text{if } k={k}'  \\
   \varepsilon         & \text{if } k\ne {k}'  \\
\end{matrix} \right.
\label{eq9}
\end{equation}
where the confidence values of non-maximum classes satisfy \( \varepsilon \rightarrow 0 \) and $K$ is the number of classes. Here \(q_j\) is a detached pseudo-target induced by the current argmax of \(p_j\) and used only for reliability analysis; it is not an externally fixed ground-truth distribution.
In practice, entropy minimization is usually implemented through a CE surrogate between the model prediction \(p_j\) and the ideal target distribution \(q_j\). We use the standard cross-entropy from the target distribution \(q_j\) to the model prediction \(p_j\), denoted \(\mathrm{CE}(q_j,p_j)=-\sum_k q_j(k)\log p_j(k)\).
The CE for a single sample can then be written as
\begin{equation}
\mathrm{CE}({{q}_{j}},{{p}_{j}})=-{{q}_{j}}({k}')\log {{p}_{j}}({k}')-\varepsilon \sum\limits_{k\ne {k}'}{\log {{p}_{j}}(k)}.
\label{eq10}
\end{equation}

For each sample $j$ and each non-maximum class $k\ne k'$, let $\delta_j(k)=p_j(k)-\mu_j$ with $\sum_{k\ne k'}\delta_j(k)=0$.
We then perform a second-order Taylor expansion of \(\log p_j(k)\) at \(p_j(k) = \mu_j\):
\begin{equation}
\log {{p}_{j}}(k)=\log {{\mu }_{j}}+\frac{{{\delta }_{j}}(k)}{{{\mu }_{j}}}-\frac{\delta _{j}^{2}(k)}{2\mu _{j}^{2}}+{{R}_{2}}({{p}_{j}}(k))
    \label{eq11}
\end{equation}
where \(R_2(p_j(k))\) is the second-order residual term.

Substituting Eq.~\ref{eq11} into Eq.~\ref{eq10} gives a compact approximation of the per-sample CE used for reliability analysis:
\begin{equation}
\begin{aligned}
    \mathrm{CE}({{q}_{j}},{{p}_{j}})&\approx -\log {{p}_{j}}({k}')+ \frac{{{(K-1)}^{3}}\varepsilon }{2{{(1-{{p}_{j}}({k}'))}^{2}}}\cdot {{v}_{j}}\\
    &+(K-1)\varepsilon \log \frac{(K-1){{p}_{j}}({k}')}{1-{{p}_{j}}({k}')}\\
    &=-{{f}_{j}({{p}_{j}}({k}'))}+{{g}_{j}({{p}_{j}}({k}'))}\cdot {{v}_{j}}
    \end{aligned}
    \label{eq12}
\end{equation}
where, for \(p=p_j(k')\), \(f_j(p)=\log p-(K-1)\varepsilon\log\frac{(K-1)p}{1-p}\) and \(g_j(p)=\frac{(K-1)^3\varepsilon}{2(1-p)^2}\).
The only non-confidence term is the residual-class variance
\begin{equation}
    {{v}_{j}}\triangleq \frac{1}{K-1}\sum\limits_{k\ne {k}'}{\delta _{j}^{2}(k)}
    \label{eq13}
\end{equation}

Eq.~\ref{eq12} yields the main selection rule: reliable pseudo-labels should have high MC and low RCV, rather than high confidence alone.
This follows because lower CE indicates higher reliability, and the CE approximation decreases as the confidence contribution \(f_j(p)\) grows while increasing with the residual-variance penalty \(g_j(p)v_j\).
With the adaptive choice \(\varepsilon=\mu_j\) introduced below, both \(f_j(p)\) and \(g_j(p)\) are increasing on \(p\in[\frac{1}{K},1)\); a concise proof is provided in the supplementary material.
Thus, larger MC lowers CE through \(-f_j(p)\), while also raising the cost of residual-class dispersion through \(g_j(p)\).
Low \(v_j\) further indicates that the remaining probability mass is not concentrated on a strong competing class.
Detailed derivations and theoretical proofs of Eqs.~\ref{eq10}--\ref{eq13} are provided in the supplementary material.

\paragraph{Setting the Ideal Residual-Class Confidence $\varepsilon$}

Although the infinitesimal residual confidence $\varepsilon$ in Eq.~\ref{eq9} is useful for analysis, training requires a concrete value, and this value should vary with the residual mass \(1-p_j(k')\), which changes across samples and training stages.
We therefore set \(\varepsilon\) to the residual mean of the model prediction,
\(\varepsilon=\mu_j=\frac{1-p_j(k')}{K-1}\).
This adaptive choice provides a usable training value while preserving the pseudo-label class and isolating the distributional shape of the non-maximum probabilities.
\begin{figure}[t]
  \centering
    \includegraphics[width=\linewidth]{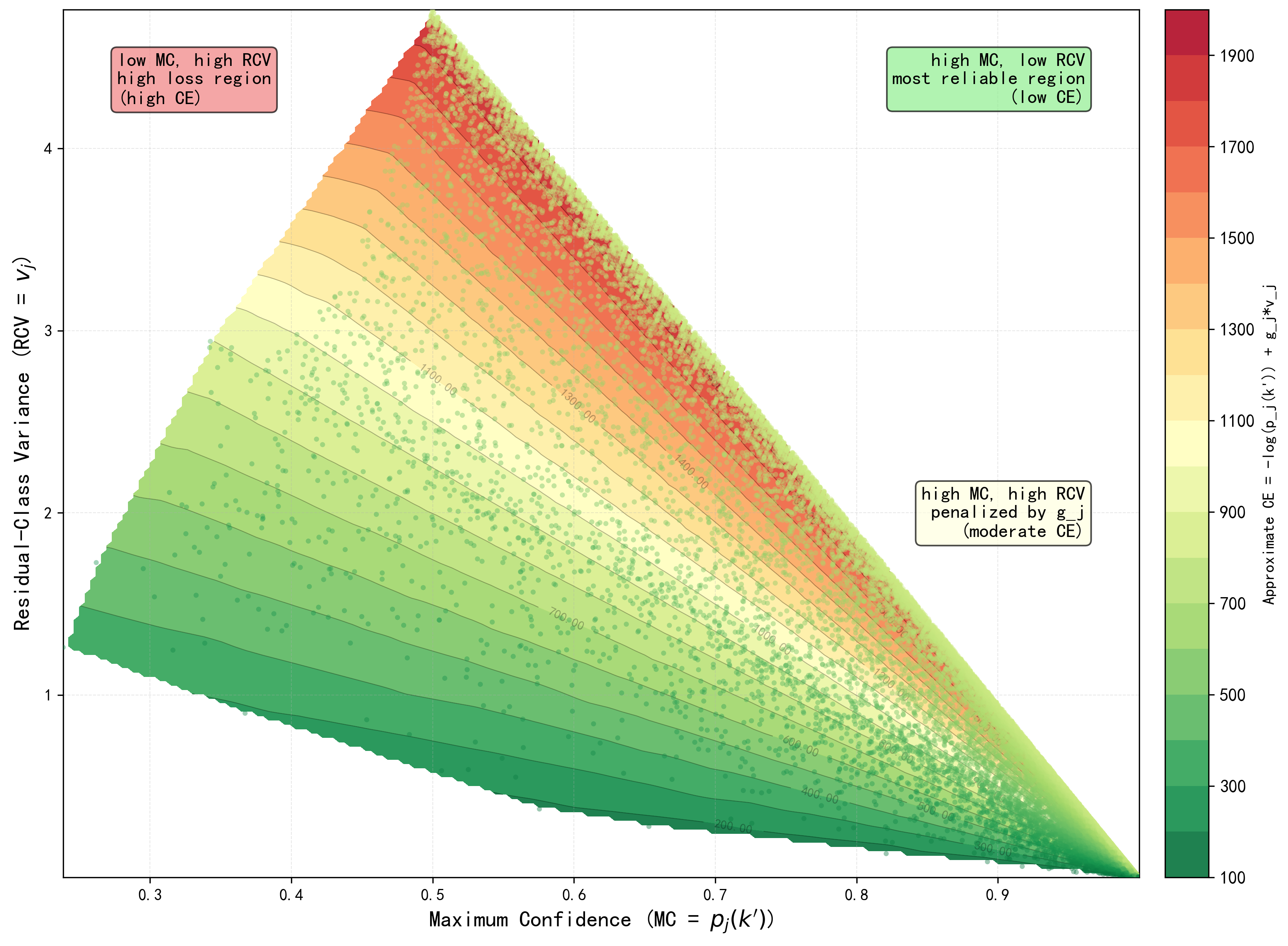}
    \caption{Illustration of the approximate CE landscape in the plane spanned by model confidence $p_j(k')$ and relative confidence variation $v_j$; see Eq.~\ref{eq15}. Low-confidence, high-variation samples incur high loss, while the most reliable pseudo-labels concentrate in the high-confidence, low-variation region.}
    \label{fig4}
\end{figure}
Substituting this choice into Eqs.~\ref{eq12}--\ref{eq13} gives
\begingroup
\setlength{\abovedisplayskip}{4pt}
\setlength{\belowdisplayskip}{4pt}
\setlength{\abovedisplayshortskip}{4pt}
\setlength{\belowdisplayshortskip}{4pt}
\begin{equation}
    \begin{aligned}
    C{{E}_{B}}
    &\approx \frac{1}{N_B}\sum_{j=1}^{N_B}\left(\mathcal{B}_j+T(p_j(k'))\right)\\
    &\ge \frac{1}{N_B}\sum_{j=1}^{N_B}\mathcal{B}_j,
    \end{aligned}    
    \label{eq15}
\end{equation}
\endgroup
where \(\mathcal{B}_j=-\log p_j(k')+\frac{(K-1)^2}{2(1-p_j(k'))}v_j\) and \(T(p_j(k')) = (1 - p_j(k'))\log \frac{(K-1)p_j(k')}{1 - p_j(k')}\).
Because \(p_j(k')=\max_k p_j(k)\), any valid probability vector satisfies \(p_j(k')\ge\frac{1}{K}\), and hence \(T(p_j(k'))\ge0\).
Dropping this non-negative term yields a valid lower bound.
Under this self-normalized form, \(v_j\) measures how far the residual-class allocation deviates from a uniform allocation with the same total residual confidence \(1-p_j(k')\) (see Fig.~\ref{fig4}); thus, the lower bound is determined entirely by the prediction $p_j$ itself.

\paragraph{Batch-Level Decomposition and Inter-Class Effects}
Eqs.~\ref{eq10}--\ref{eq13} show that ``high MC + low RCV'' is a useful local indicator of low approximate CE loss, but applying this rule independently to each sample in a mini-batch remains limited.
The approximation is local and depends on the ideal target distribution and higher-order residual terms, whereas practical mini-batches are drawn from long-tailed data and contain class-dependent confidence distributions with imbalanced class counts.
As a result, per-sample ranking or truncation can behave like a refined threshold rule and inherit majority-class selection bias.
We therefore analyze the batch objective directly and rewrite the average batch CE as the three-term decomposition in Eq.~\ref{eq14}:
\begin{equation}
    \begin{aligned}
    CE_B
    &\approx -\frac{1}{N_B}\sum_{j=1}^{N_B} f_j
      +\frac{1}{N_B}\sum_{j=1}^{N_B}(g_j\cdot v_j)\\
    &=-\underbrace{\frac{1}{N_B}\sum_{j=1}^{N_B} f_j}_{\bar{C}}
      +\underbrace{\bar{g}\cdot \bar{v}}_{\mathrm{sRCV}}
      +\mathrm{Cov}(g,v)
    \end{aligned}
    \label{eq14}
\end{equation}
where \( \bar{C}=\frac{1}{N_B}\sum_j f_j \), \( \mathrm{sRCV}=\bar{g}\bar{v} \), \( \bar{g} = \frac{1}{N_B} \sum_j g_j({{p}_{j}}({k}')) \), \( \bar{v} = \frac{1}{N_B} \sum_j v_j \), and \( \mathrm{Cov}(g, v) = \frac{1}{N_B} \sum_j (g_j({{p}_{j}}({k}')) - \bar{g})(v_j - \bar{v}) \).

Eq.~\ref{eq14} gives three batch-level quantities with clear desirable directions: a larger confidence term $\bar{C}$ lowers the batch CE, a smaller scaled variance term $\mathrm{sRCV}=\bar{g}\bar{v}$ suppresses residual-class competition, and a smaller $\mathrm{Cov}(g,v)$ indicates weaker coupling between high confidence and large residual variance within the batch.
Thus, reliable predictions should concentrate in the high-MC and low-weighted-RCV region, whereas samples affected by overconfidence or class imbalance are pushed away from this region.
CoVar uses this batch structure to motivate the following prediction-separation module: by partitioning samples in the two-dimensional space $[\log p_j(k'), -g_j(p_j(k'))v_j]$, the method reduces majority-class dominance and improves head-tail balance without an explicit class-frequency reweighting rule.

\subsection{Prediction Separation via Spectral Relaxation}

The second-order CE approximation shows that prediction reliability depends jointly on the maximum confidence $p_j(k')$ and residual-class variance $v_j$, with a confidence-dependent weight on $v_j$.
Because this nonlinear confidence--variance coupling cannot be captured by a fixed threshold on confidence alone, we represent each prediction with a confidence--variance embedding and reformulate pseudo-label selection as a partitioning problem that separates high- and low-reliability predictions in this feature space.

To obtain a tractable solution, we adopt a spectral relaxation of the discrete partitioning problem. 
This yields a continuous optimization formulation that captures the interaction between confidence and variance and enables adaptive, data-driven separation of reliable predictions (Fig.~\ref{fig3}).

\paragraph{Attribute Embedding and Problem Formulation}

Following Eq.~\ref{eq12}, each prediction is represented by a reliability-aware embedding composed of the confidence term and the weighted RCV term:
\begin{equation}
{{\tilde{h}}_{j}}=
\left[
\log {{p}_{j}}({k}'),\;
-\frac{{{(K-1)}^{2}}}{2(1-{{p}_{j}}({k}'))}\cdot {{v}_{j}}
\right]^{T}.
    \label{eq17}
\end{equation}

Let \( \Phi = [\mathbf{h}_1, \mathbf{h}_2, \ldots, \mathbf{h}_N] \in \mathbb{R}^{2 \times N} \) collect all embeddings, and let $X = \Phi^T \in \mathbb{R}^{N \times 2}$ be the sample-wise representation.
The inner-product structure in this reliability space enables spectral methods to separate predictions into high- and low-reliability subsets without manually tuned confidence thresholds.

Our goal is to partition these embeddings into two clusters such that predictions with both high MC and low RCV are grouped together.
Following the normalized partition view discussed in the supplementary material, we optimize the normalized spectral objective:
\begin{equation}
\begin{aligned}
\underset{\mathbf{S}}{\mathop{\max }}\,
&\operatorname{Tr}\left( (S^T S)^{-1/2} S^T X X^T S (S^T S)^{-1/2} \right),\\
&\text{s}\text{.t}\text{. }S\in {{\{0,1\}}^{N\times 2}}
\end{aligned}
    \label{eq18}
\end{equation}
where \( \operatorname{Tr}(\cdot) \) denotes the matrix trace, $\sum\limits_{c}{{{S}_{n,c}}=1}$, and \( S \) is the two-way reliability assignment matrix constrained so that each row has exactly one nonzero entry equal to 1. The factor \((S^T S)^{-1/2}\) normalizes for cluster size.

\paragraph{Spectral Relaxation and Discretization}

The combinatorial problem is NP-hard, so we follow spectral clustering~\citep{relaxation} by setting \(Y=S(S^T S)^{-1/2}\) and relaxing the discrete indicator constraint to \(Y^T Y=I\).
By the Ky Fan theorem~\citep{kf}, the relaxed solution is given by the first two eigenvectors of \(XX^T\).
Since each embedding in Eq.~\ref{eq17} is two-dimensional, \(\Phi\in\mathbb{R}^{2\times N}\) is a thin matrix. Therefore, instead of forming the pixel-level \(N\times N\) Gram matrix, we compute the economy SVD of \(\Phi\):
\begin{equation}
    \Phi = U\Sigma V^T, \quad U\in\mathbb{R}^{2\times 2},\; \Sigma\in\mathbb{R}^{2\times 2},\; V\in\mathbb{R}^{N\times 2}.
    \label{eq:svd}
\end{equation}
The columns of \(V\) provide the relaxed spectral coordinates used for assignment.
For discretization, we use a simple deterministic rounding rule on the two-dimensional relaxed spectral embedding:
\begin{equation}
S_{n,c}^{*}=\Delta (\arg {{\max }_{i\in \{1,2\}}}|\mathbf{v}_{i}(n)|,\ c)
    \label{eq19}
\end{equation}
where \( \mathbf{v}_1, \mathbf{v}_2 \) are the right singular vectors of \( \Phi \), \( |\mathbf{v}_c(j)| \) denotes the absolute value of the \( j \)-th element of \( \mathbf{v}_c \), and \( \Delta(\cdot, \cdot) \) is the Kronecker delta function.
This assigns each prediction to the cluster corresponding to its dominant relaxed spectral component; we refer to this procedure as SVD-based spectral relaxation.

\paragraph{Projection and Weight Construction}

Based on the batch-level criterion in Eq.~\ref{eq14}, high-reliability predictions should have high MC and low weighted RCV in the embedding space.
We therefore identify the high-reliability cluster $c^*$ by comparing the centroids of the two candidate clusters.
For each candidate cluster $c \in \{1, 2\}$, we compute
\begin{equation}
    \bar{h}_c = \frac{\sum_{n:\,S_{n,c}^*=1} \tilde{h}_n}{\sum_{n} S_{n,c}^*},
    \label{eq:centroid}
\end{equation}
where $\tilde{h}_n = [\tilde{h}_n(1),\, \tilde{h}_n(2)]^T$ as defined in Eq.~\ref{eq17}, with $\tilde{h}_n(1)=\log p_n(k')$ and $\tilde{h}_n(2) = -\frac{(K-1)^2}{2(1-p_n(k'))}\cdot v_n$.
The high-reliability cluster is then selected as
\begin{equation}
    c^* = \arg\max_{c \in \{1,2\}} \left[\bar{h}_c(1) + \bar{h}_c(2)\right],
    \label{eq:cluster-select}
\end{equation}
i.e., the cluster whose centroid has higher mean log MC and higher mean negated weighted RCV, or equivalently lower mean weighted RCV.
This directly follows the high-MC/low-RCV reliability criterion in Eq.~\ref{eq14}.

We define a projection matrix \( Z = \Phi S_{:,c^*}^* \), which maps predictions to their reliable components.
Smooth loss weights are then constructed with a Gaussian weighting function:
\begin{equation}
{{\omega }_{n}}=\prod\limits_{d=1}^{2}{\exp }\left( \frac{{{({{{\tilde{h}}}_{n}}(d)-{{\mu }_{c^*,d}})}^{2}}}{-2 \sigma _{c^*,d}^{2}} \right)
    \label{eq20}
\end{equation}
where \( \mu_{c^*,d} \) and \( \sigma_{c^*,d} \) are the mean and standard deviation of the predictions assigned to the high-reliability cluster \(c^*\) along embedding dimension \( d \in \{1,2\} \).
For predictions assigned to the high-reliability cluster and lying above the high-reliability centroid in both embedding coordinates, the weight is fixed to 1.
Finally, after the high-reliability cluster \(c^*\) is identified, define the high-confidence side of the reliable cluster as \(\mathcal{R}_{c^*}\). The task-dependent indicator mapping \( M_i \) is:
\begin{equation}
  \begin{gathered}
  \mathcal{R}_{c^*}=\{n:\ S_{n,c^*}^{*}=1,\ 
  \tilde{h}_n(d)>\mu_{c^*,d},\ d=1,2\},\\
  {{M}_{i}}(n)=
  \begin{cases}
  1, & n\in\mathcal{R}_{c^*},\\
  {{\omega }_{n}}, & \text{otherwise}.
  \end{cases}
  \end{gathered}
  \label{eq21}
\end{equation}
This method distinguishes reliable from unreliable predictions based on the joint feature distribution of confidence and dispersion.
It is general and applicable to arbitrary classification tasks.

\section{Experiments}

\subsection{Datasets and Experimental Setup}

We evaluate the proposed method on two representative tasks: semi-supervised semantic segmentation and semi-supervised image classification. 

\textbf{Semantic segmentation.}
For semantic segmentation, experiments are conducted on two benchmark datasets, VOC augmented with SBD and Cityscapes~\citep{pascal,cityscapes}. 
PASCAL VOC~2012 is a standard dataset containing 20 object categories and one background class, with 1,464 training images and 1,449 validation images. 
Following common practice, the training set was augmented with 9,118 additional images from the Segmentation Boundary Dataset~\citep{SBD}, resulting in a total of 10,582 training images in VOC augmented with SBD. 
Cityscapes focuses on urban scene understanding and includes 19 categories, comprising 2,975 finely annotated training images and 500 validation images at a resolution of $1024\times2048$.

\textbf{Image classification.}
For semi-supervised image classification, we use four standard benchmarks: CIFAR-10, CIFAR-100, SVHN, and STL-10~\citep{krizhevsky2009learning,netzer2011reading,coates2011stl10}.
CIFAR-10 consists of 10 classes with 60,000 color images of size $32\times32$ (50,000 for training and 10,000 for testing).
CIFAR-100 shares the same image format as CIFAR-10 but contains 100 fine-grained classes, providing a more challenging setting with richer inter-class residual-distribution heterogeneity.
SVHN is a real-world digit recognition dataset containing over 600,000 images of size $32\times32$ obtained from house numbers in Google Street View, spanning 10 classes; its high within-class variation and heavy class imbalance make it a demanding benchmark for pseudo-label selection.
STL-10 contains 10 classes with 5,000 labeled images ($96\times96$) and 100,000 unlabeled images drawn from a broader distribution, intentionally mismatched from the labeled set to stress-test generalization under label scarcity.
Following the standard semi-supervised protocol, we evaluate CIFAR-10 with 4, 10, and 100 labeled samples per class, and CIFAR-100, SVHN, and STL-10 with 4, 10, and 25 labeled samples per class; the remaining training samples are treated as unlabeled.

\begin{table*}[t]
    \centering
  \caption{Comparison with prior work on VOC augmented with SBD. The 1/16, 1/8, and 1/4 settings refer to labeled subsets drawn from the 10,582-image training pool. Entries marked $^\dag$ follow the official U2PL splits; other entries are reported under the protocols of the respective methods. Backbone and crop size are listed per row. All mean$\pm$sd values are computed over three random seeds. Boldface and underlined values denote the best and second-best results, respectively.}
    \setlength{\tabcolsep}{2mm}{
    \begin{tabular}{cc|cccc}
    \toprule
    VOC augmented with SBD (mIoU, \%)  & Backbone   & Crop Size & 1/16    & 1/8   & 1/4   \\
    \midrule
    ST++~\citep{ST++}           & ResNet-101 & 321$\times$321  & {74.5}   & 76.3   & 76.6   \\
    UniMatch V1~\citep{UniMatch}   & ResNet-101 & 321$\times$321  & {76.5}   & 77.0   & 77.2   \\
    CorrMatch~\citep{CorrMatch} & ResNet-101 & 321$\times$321  & {77.6}   & {77.8} & {78.3} \\
    CAC~\citep{CAC}             & ResNet-101 & 321$\times$321  & {72.4}   & 74.6   & 76.3   \\
    CSL~\citep{Liu_2025_ICCV}    & ResNet-101 & 321$\times$321  & {\underline{77.8}} & \underline{78.5} & \underline{79.0} \\
    CoVar (Ours)                & ResNet-101 & 321$\times$321  & {\textbf{78.6}{\tiny ±0.31}} & \textbf{78.9}{\tiny ±0.22} & \textbf{79.8}{\tiny ±0.40} \\
    \midrule
    ST++~\citep{ST++}    & ResNet-101 & 513$\times$513 & {74.7} & 77.9  & 77.9 \\
    UniMatch V1~\citep{UniMatch}  &  CLIP-B & 513$\times$513 & {78.1} & 78.4  & 79.2 \\
    CorrMatch~\citep{CorrMatch}  & ResNet-101 & 513$\times$513 & {78.4} & 79.3  & 79.6 \\
    ESL~\citep{ESL}     & ResNet-101 & 513$\times$513 & {76.4} & 78.6  & 79.0 \\
    PS-MT~\citep{PS-MT}     & ResNet-101  & 513$\times$513 & {75.5} & 78.2  & 78.7 \\
    CFCG~\citep{CFCG}   & ResNet-101 & 513$\times$513 & {76.8} & 79.1  & 80.0 \\
    CCVC~\citep{CCVC}    & ResNet-101  & 513$\times$513 & {77.2} & 78.4  & 79.0 \\
    DLG~\citep{DLG}      & ResNet-101 & 513$\times$513 & {77.8} & 79.3  & 79.1 \\
    RankMatch~\citep{rankmatch} & ResNet-101 & 513$\times$513 & {78.9} & 79.2    & 80.0 \\
    DGCL~\citep{DGCL}     & ResNet-101 & 513$\times$513 & {76.6} & 78.3  & 79.3 \\
    AllSpark~\citep{allspark}  &  MiT-B5 & 513$\times$513 & {78.3} & {80.0}  & {80.4} \\
    DDFP~\citep{DDFP} & ResNet-101 & 513$\times$513 & {78.3} & 78.9    & 79.8 \\
    CSL~\citep{Liu_2025_ICCV}      & ResNet-101 & 513$\times$513 & {78.9} & {79.9} & {80.3} \\
    UniMatch V2~\citep{Unimatchv2}      & DINOv2-B & 513$\times$513 & \underline{85.3} & \underline{87.9 } & \underline{88.8} \\
    CoVar (Ours)       & ResNet-101 & 513$\times$513 & {79.7{\tiny ±0.29}} & {80.4{\tiny ±0.41}} & {80.8{\tiny ±0.18}} \\
    CoVar (Ours)       & DINOv2-B & 513$\times$513 & {\textbf{86.6}{\tiny ±0.36}} & \textbf{89.1}{\tiny ±0.45} & \textbf{89.9}{\tiny ±0.48} \\
    \midrule
    U2PL$^\dag$~\citep{U2PL}      & ResNet-101 & 513$\times$513 & {77.2} & 79.0  & 79.3 \\
    GTA$^\dag$~\citep{GTA}     & ResNet-101 & 513$\times$513 & {77.8} & 80.4  & 80.5 \\
    CorrMatch$^\dag$~\citep{CorrMatch} & ResNet-101 & 513$\times$513 & {81.3} & 80.9  & {81.9} \\
    UniMatch V1$^\dag$~\citep{UniMatch} & CLIP-B & 513$\times$513 & {81.0} & 80.4  & 81.9 \\
    AugSeg$^\dag$~\citep{augseg}  & ResNet-101 & 513$\times$513 & {79.3} & 80.5  & 81.5 \\
    AllSpark$^\dag$~\citep{allspark} & MiT-B5 & 513$\times$513 & \underline{81.6} & {80.9}  & 82.0 \\
    CSL$^\dag$~\citep{Liu_2025_ICCV}      & ResNet-101 & 513$\times$513 & {81.6} & \underline{81.1} & \underline{82.4} \\
    CoVar (Ours)$^\dag$       & ResNet-101 & 513$\times$513 & {\textbf{81.9}{\tiny ±0.19}} & \textbf{82.3}{\tiny ±0.23} & \textbf{82.8}{\tiny ±0.37} \\
    \bottomrule
    \end{tabular}
    }
    \label{tab:1}
\end{table*} 

\textbf{Implementation Details.}
For implementation, CoVar is inserted into the UniMatch training pipeline with a ResNet-101 backbone for semi-supervised semantic segmentation, and we additionally evaluate compatibility with the DINOv2-B setting used by UniMatch~V2. Other published methods are included as reference baselines, with backbone and crop size specified in Tables~\ref{tab:1} and \ref{tab:2}.  
All experiments use stochastic gradient descent with a batch size of 8 and a weight decay of \(1 \times 10^{-4}\).  
For PASCAL VOC 2012, the initial learning rate is 0.001, crop sizes are set to $321\times321$ or $513\times513$, and training is performed for 80 epochs.  
The decoder learning rate is set to 10 times that of the backbone.  
For Cityscapes, we use the same initial learning rate with a crop size of $801\times801$, train for 240 epochs, and adopt the Online Hard Example Mining loss for fair comparison.

For semi-supervised image classification, we build on SimPLE with a Wide ResNet-28-2 backbone and follow the standard protocols on CIFAR-10, CIFAR-100, SVHN, and STL-10. Weak augmentation uses crop and flip perturbations with normalization, while strong augmentation adds RandAugment-based transformations. We optimize with SGD using Nesterov momentum and cosine decay, and insert CoVar after pseudo-label guessing and before computing the unsupervised and pairwise losses. Dataset-specific thresholds inherited from the SimPLE baseline, loss coefficients, augmentation-view settings, and remaining implementation details are provided in the supplementary material.
\begin{table*}[t]
    \centering
    \caption{Comparison with state-of-the-art methods on Cityscapes.}
    \setlength{\tabcolsep}{2.9mm}{
    \begin{tabular}{cc|cccc}
    \toprule
    Cityscapes (mIoU, \%) & Backbone & 1/16(186) & 1/8(372) & 1/4(744) & 1/2(1488) \\
    \midrule
    Supervised & ResNet-101 & {63.1} & {70.5} & {73.1} & {76.2} \\
    ST++~\citep{ST++}   & ResNet-101 & {67.6} & {73.4} & {74.6} & {77.8} \\
    UniMatch V1~\citep{UniMatch} & CLIP-B & {76.6} & {77.9} & {79.2} & {79.5} \\
    CorrMatch~\citep{CorrMatch} & ResNet-101 & {77.3} & {78.5} & {79.4} & {80.4} \\
    ESL ~\citep{ESL}    & ResNet-101 & {75.1} & {77.2} & {79.0} & {80.5} \\
    DGCL~\citep{DGCL}   & ResNet-101 & {73.2} & {77.3} & {78.5} & {80.7} \\
    AugSeg~\citep{augseg}  & ResNet-101 & {75.2} & {77.8} & {79.6} & {80.4} \\
    CCVC~\citep{CCVC}   & ResNet-101 & {74.9} & {76.4} & {77.3} & {-} \\
    DAW~\citep{DAW}    & ResNet-101 & {76.6} & {78.4} & {79.8} & {80.6} \\
    U2PL~\citep{U2PL}    & ResNet-101  & {70.3} & {74.4} & {76.5} & {79.1} \\
    CPS~\citep{cps}   & ResNet-101 & {69.8} & {74.3} & {74.6} & {76.8} \\
    DDFP~\citep{DDFP} & ResNet-101 & 77.1 & 78.2 & {79.9} & {80.8} \\
    SemiVL~\citep{hoyer2024semivl} & CLIP-B & {72.2} & {75.4} & {77.2} & {79.6} \\
    FPL~\citep{FPL}   & ResNet-101 & {75.7} & {78.5} & {79.2} & {-} \\
    AllSpark~\citep{allspark}  &  MiT-B5 & {78.3} & {79.2}  & {80.6}  & {81.4}\\ 
    CSL~\citep{Liu_2025_ICCV}      & ResNet-101 & {78.2} & {78.8} & {80.0} & {81.1} \\
    UniMatch V2~\citep{Unimatchv2}      & DINOv2-B & \underline{82.6} & \underline{83.3} & \underline{83.5} & \underline{84.1} \\
    CoVar (Ours)      & ResNet-101  & {78.7{\tiny ±0.28}} & {79.3{\tiny ±0.24}} & {80.6{\tiny ±0.30}} & {81.5{\tiny ±0.22}} \\
    CoVar (Ours)      & DINOv2-B  & \textbf{83.7}{\tiny ±0.39} & {\textbf{84.4}{\tiny ±0.43}} & \textbf{85.0}{\tiny ±0.36} & \textbf{85.3}{\tiny ±0.37} \\
    \bottomrule
    \end{tabular}
    }
    \label{tab:2}
\end{table*} 

\textbf{Evaluation Metrics.}
For evaluation, we report mean Intersection over Union (mIoU) on the standard validation sets of PASCAL VOC 2012 and Cityscapes for semantic segmentation; Cityscapes is evaluated with sliding windows of size $801\times801$.

For semi-supervised image classification, we report Top-1 error rate (\%, lower is better) on the standard test sets of CIFAR-10, CIFAR-100, SVHN, and STL-10.
Ablation studies use VOC augmented with SBD at $321\times321$ crops and CIFAR-10 unless otherwise specified.

\subsection{Performance Improvement When Integrated with Existing State-of-the-Art Methods}
To validate the versatility and effectiveness of the proposed framework, we evaluate it on two representative semi-supervised tasks: semantic segmentation and image classification.
For semantic segmentation, we integrate CoVar into the UniMatch pipeline under ResNet-101 and additionally test it in the DINOv2-B setting of UniMatch~V2; for image classification, we build on SimPLE and insert CoVar before the unsupervised and pairwise losses.
For each setting, we compare against the corresponding baseline under multiple label partitions to assess performance changes across label-scarce regimes.
\begin{table*}[htbp]
  \centering
  \small
  \caption{Top-1 error rate (\%) on CIFAR-10, CIFAR-100, SVHN, and STL-10 (lower is better).}
  \setlength{\tabcolsep}{0.4mm}{
  \begin{tabular}{l *{12}{c}}
  \toprule
  & \multicolumn{3}{c}{\textbf{CIFAR-10}} & \multicolumn{3}{c}{\textbf{CIFAR-100}} & \multicolumn{3}{c}{\textbf{SVHN}} & \multicolumn{3}{c}{\textbf{STL-10}} \\
  \cmidrule(lr){2-4}\cmidrule(lr){5-7}\cmidrule(lr){8-10}\cmidrule(lr){11-13}
  \# Labels/Class & 4 & 10 & 100 & 4 & 10 & 25 & 4 & 10 & 25 & 4 & 10 & 25 \\
  \midrule
  PseudoLabel & 76.25 & 65.23 & 48.22 & 87.09 & 76.02 & 59.04 & 75.89 & 27.54 & 16.66 & 73.75 & 68.12 & 49.26 \\
  MeanTeacher & 76.86 & 55.84 & 57.10 & 90.40 & 64.52 & 61.11 & 81.98 & 13.84 & 25.16 & 70.90 & 64.42 & 44.44 \\
  MixMatch~\citep{mixmatch} & 70.62 & 53.65 & 37.32 & 80.00 & 64.37 & 49.52 & 79.57 & 24.74 & 3.76 & 68.72 & 55.70 & 42.28 \\
  ReMixMatch~\citep{berthelot2019remixmatch} & 14.55 & 11.15 & 9.26 & 57.15 & 42.83 & 34.72 & 31.33 & 6.22 & 6.43 & 46.61 & 30.20 & 21.45 \\
  FixMatch~\citep{fixmatch} & 8.27 & 7.75 & 5.02 & 53.43 & 42.69 & 34.23 & 3.62 & \textbf{2.05} & \underline{2.08} & 40.74 & 16.47 & 9.75 \\
  FlexMatch~\citep{flexmatch} & 5.14 & 5.31 & 4.96 & 50.09 & 39.42 & 33.39 & 3.43 & 3.63 & 4.95 & 22.19 & \underline{12.20} & \underline{8.67} \\
  SoftMatch~\citep{softmatch} & 5.13 & 5.03 & {4.91} & 49.58 & 39.00 & 33.10 & 2.92 & 2.25 & {2.09} & 22.40 & \textbf{11.55} & \textbf{7.96} \\
  FreeMatch~\citep{freematch} & \underline{4.90} & \underline{4.95} & 4.91 & 49.18 & 39.23 & 32.85 & 3.75 & 4.23 & 4.14 & 26.52 & 12.81 & 8.84 \\
  FullFlex~\citep{chen2023boosting} & 6.16 & 5.01 & {4.93} & \underline{48.40} & 38.81 & 32.81 & {2.77} & 3.67 & 4.68 & 33.35 & 14.70 & 8.88 \\
  CGMatch~\citep{cheng2025cgmatch} & {4.92} & {4.96} & \underline{4.85} & \textbf{47.49} & \textbf{38.62} & \underline{32.47} & \underline{2.45} & {2.14} & \textbf{2.06} & \underline{22.13} & 16.18 & 9.42 \\
  SimPLE~\citep{hu2021simple} & {5.89} & {5.40} & {5.13} & {50.62} & {42.17} & {35.41} & {3.83} & {3.13} & {2.25} & {41.56} & 16.94 & 9.91 \\
  CoVar (Ours) & \textbf{4.81} & \textbf{4.71} & \textbf{4.75} & \textbf{47.49} & \underline{38.75} & \textbf{32.44} & \textbf{2.34} & \underline{2.12} & \underline{2.08} & \textbf{22.12} & 12.37 & 8.74 \\
               & {\tiny ±0.29} & {\tiny ±0.33} & {\tiny ±0.27} & {\tiny ±0.21}  & {\tiny ±0.30}     & {\tiny ±0.20}  & {\tiny ±0.19} & {\tiny ±0.22}    & {\tiny ±0.15}    & {\tiny ±0.23}  & {\tiny ±0.32} & {\tiny ±0.18} \\
  \bottomrule
  \end{tabular}
  }
  \label{tab3}
\end{table*}

\textbf{Semi-supervised semantic segmentation.}
For semi-supervised semantic segmentation, the core like-for-like comparisons on VOC augmented with SBD and Cityscapes are the ResNet-101 rows in Tables~\ref{tab:1} and \ref{tab:2}. Under this matched backbone, CoVar consistently improves over CorrMatch and CSL across all reported splits.
On VOC with $321\times321$ crops, CoVar reaches 78.6 / 78.9 / 79.8 mIoU at the 1/16 / 1/8 / 1/4 splits, improving CorrMatch by +1.0 / +1.1 / +1.5 and CSL by +0.8 / +0.4 / +0.8. On VOC with $513\times513$ crops, it reaches 79.7 / 80.4 / 80.8 mIoU, exceeding CorrMatch by +1.3 / +1.1 / +1.2 and CSL by +0.8 / +0.5 / +0.5 across the 1/16 / 1/8 / 1/4 splits. Under the U2PL protocol, CoVar\textsuperscript{\dag} (81.9 / 82.3 / 82.8) improves CSL\textsuperscript{\dag} (81.6 / 81.1 / 82.4) by +0.3 / +1.2 / +0.4, indicating that the adaptive coupling of MC and RCV remains beneficial under a different training recipe.

On Cityscapes (Table~\ref{tab:2}), CoVar with ResNet-101 obtains 78.7 / 79.3 / 80.6 / 81.5 mIoU at the 1/16 / 1/8 / 1/4 / 1/2 splits, improving CorrMatch by +1.4 / +0.8 / +1.2 / +1.1 and outperforming CSL by +0.5 / +0.5 / +0.6 / +0.4.
Separately, when paired with the stronger DINOv2-B setting used by UniMatch~V2, CoVar improves by +1.1 / +1.1 / +1.5 / +1.2 mIoU on Cityscapes and by +1.3 / +1.2 / +1.1 mIoU on VOC $513\times513$. The CLIP-B and MiT-B5 rows in Tables~\ref{tab:1}--\ref{tab:2} are therefore treated as reference baselines rather than strict matched-backbone comparisons.

\textbf{Semi-supervised image classification.}
For semi-supervised image classification, CoVar achieves the best error rate on CIFAR-10 across all three label regimes (4.81 / 4.71 / 4.75 at 4 / 10 / 100 labels per class; Table~\ref{tab3}), outperforming the strongest prior entry in each regime by 0.09 / 0.24 / 0.10 and reducing the error of SimPLE by 1.08 / 0.69 / 0.38. These results confirm that residual dispersion filtering yields consistent gains even near saturation.
On CIFAR-100, where richer class structure amplifies residual-distribution heterogeneity, CoVar matches CGMatch at 4 labeled samples per class (47.49), ranks second at 10 labeled samples per class (38.75 compared with 38.62), and achieves the best error at 25 labeled samples per class (32.44 compared with 32.47), while reducing the error of SimPLE by 3.13 / 3.42 / 2.97 at 4 / 10 / 25 labeled samples per class.
On SVHN, CoVar ranks first at 4 labels/class (2.34, $-$1.49 over SimPLE) and second at 10 and 25 labels/class (2.12 / 2.08). On STL-10, CoVar leads at 4 labels/class (22.12) and remains competitive at 10 and 25 labels/class, where the margin becomes smaller at higher label counts.
This mirrors segmentation: improvements stem from discriminating high-confidence predictions that still contain a strong competing non-maximum class rather than replacing simple thresholds.

\begin{figure}[t]
  \centering
    \includegraphics[width=\linewidth]{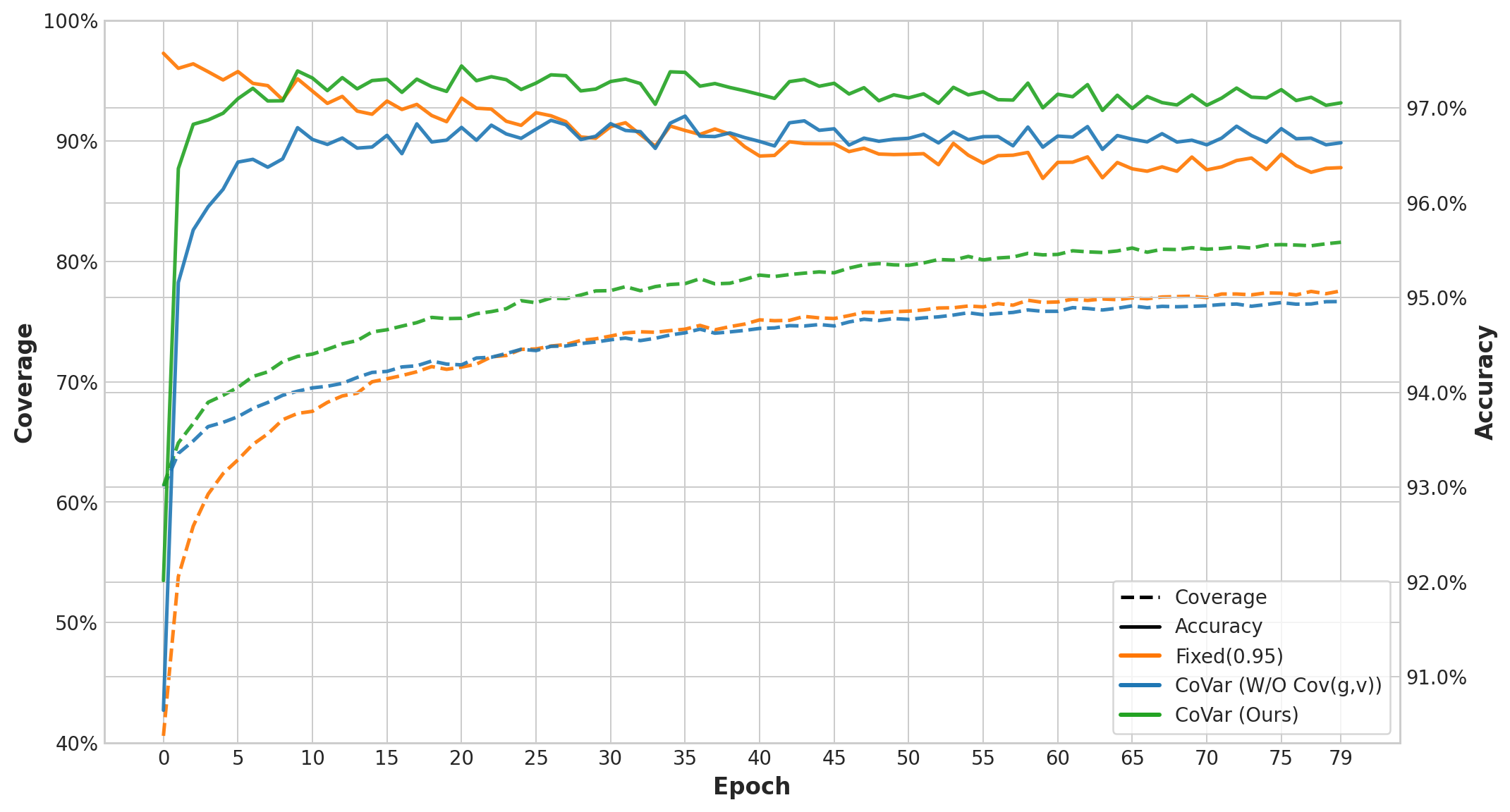}
    \caption{Changes in pseudo-label coverage during training.
    Experiments are conducted on VOC augmented with SBD (321$\times$321).
    Compared with fixed-threshold selection, CoVar improves pseudo-label coverage while maintaining high accuracy.}
    \label{fig:selrate}
\end{figure}

\begin{figure*}[t]
  \centering
  \begin{minipage}[t]{0.24\linewidth}
    \centering
    \includegraphics[width=\linewidth]{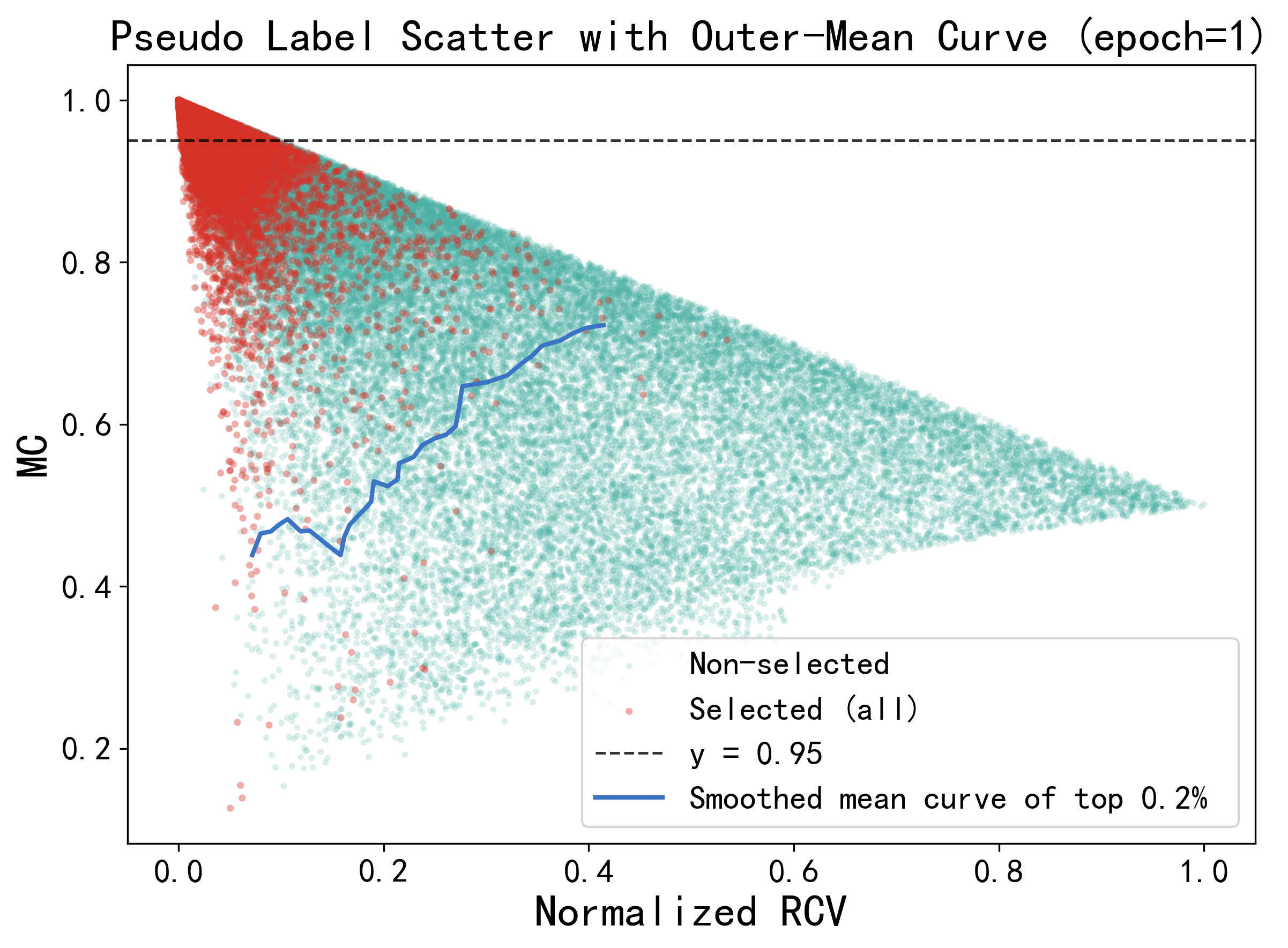}
  \end{minipage}\hfill
  \begin{minipage}[t]{0.24\linewidth}
    \centering
    \includegraphics[width=\linewidth]{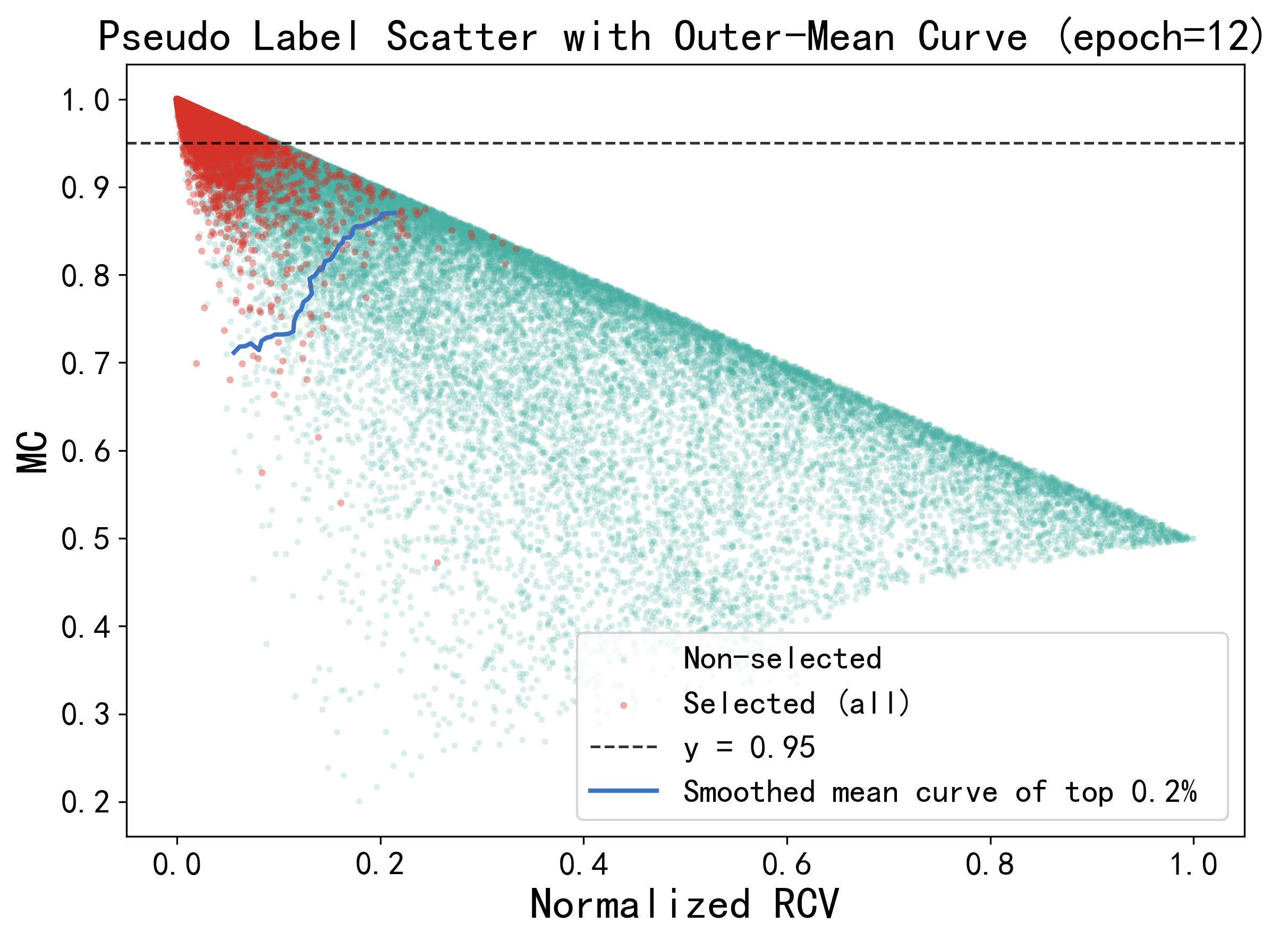}
  \end{minipage}\hfill
  \begin{minipage}[t]{0.24\linewidth}
    \centering
    \includegraphics[width=\linewidth]{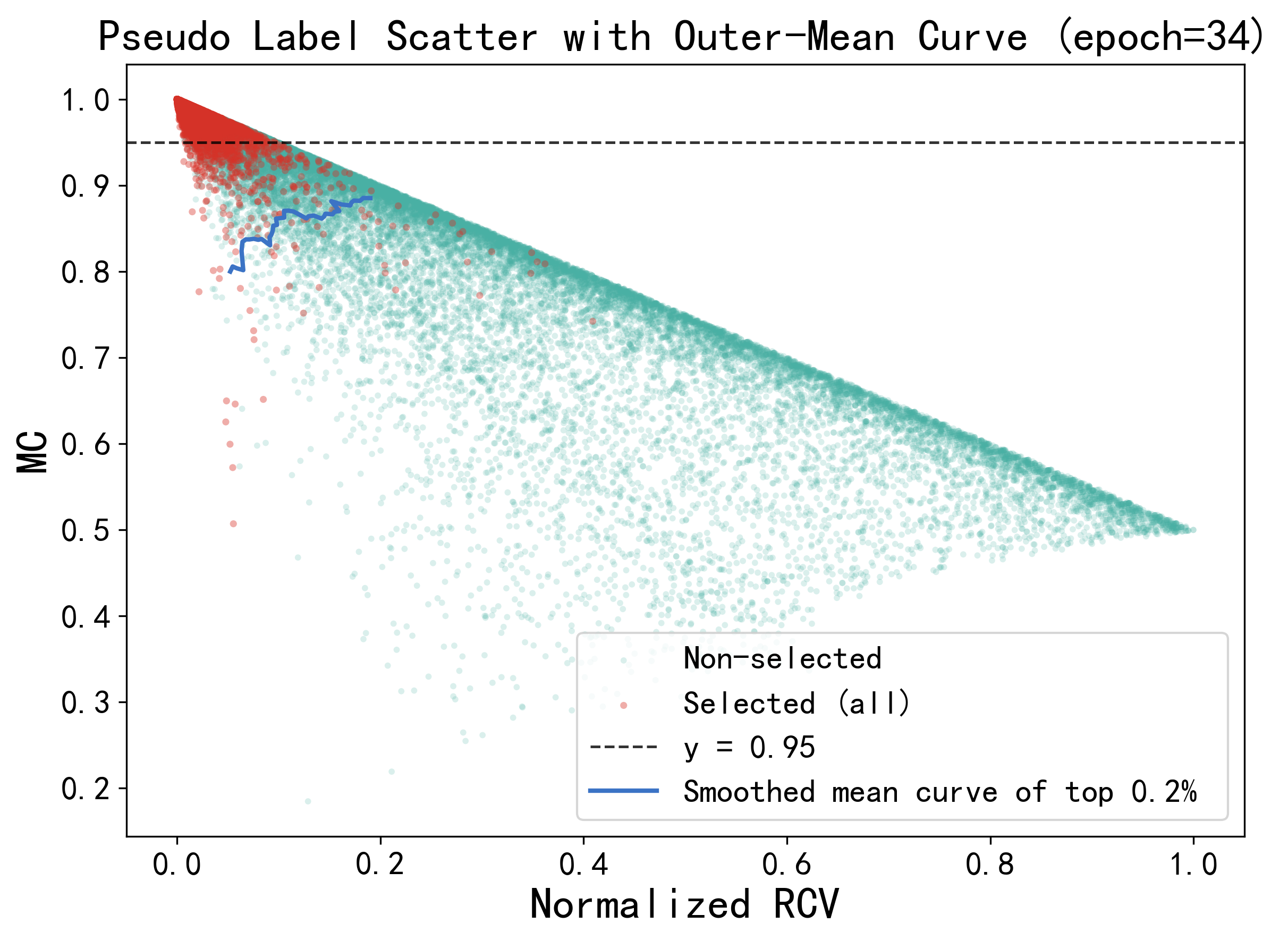}
  \end{minipage}\hfill
  \begin{minipage}[t]{0.24\linewidth}
    \centering
    \includegraphics[width=\linewidth]{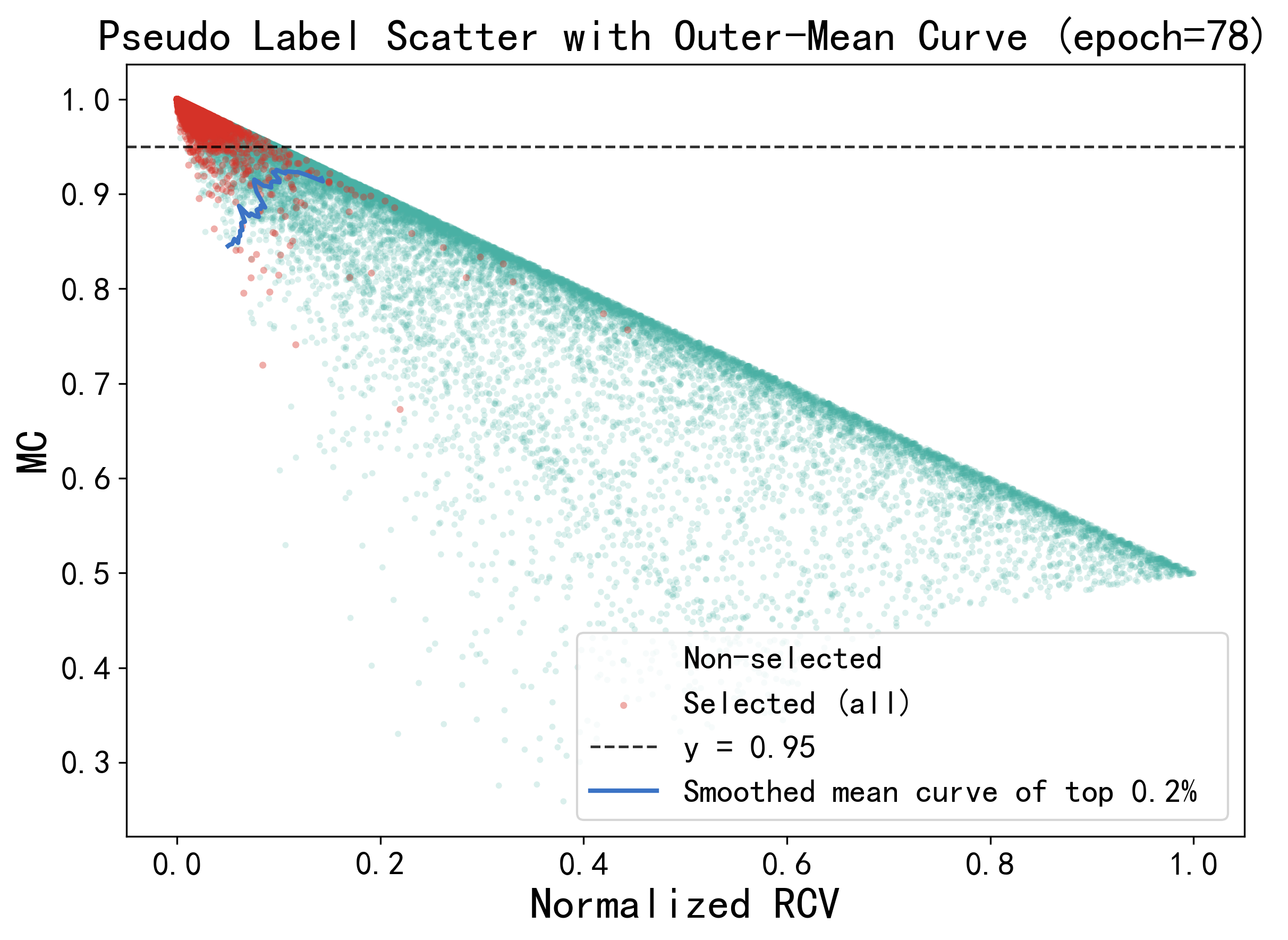}
  \end{minipage}
  \caption{Variation of sample-level MC and RCV during training. 
  Experiments are conducted on VOC augmented with SBD (1/4, 321$\times$321, 100,000 sample points are randomly sampled in each epoch).
  The fixed-threshold method selects only those sample points with $MC > 0.95$.
  As training progresses, CoVar tends to select reliable pseudo-labels with high MC and low RCV in the remaining classes, rather than relying solely on MC.}
  \label{fig:mc-rcv-evolution}
\end{figure*}

\begin{figure}[t]
  \centering
    \includegraphics[width=\linewidth]{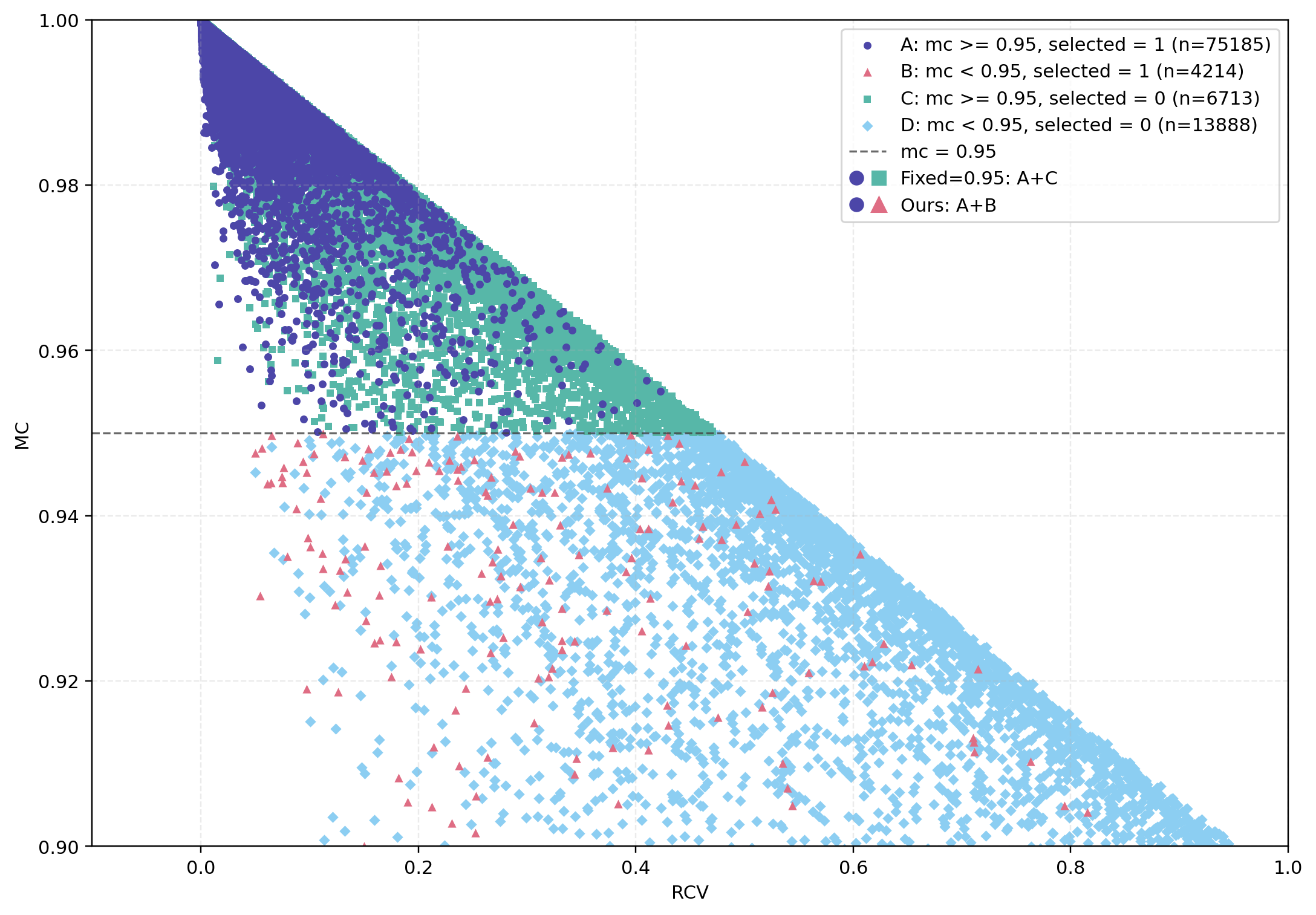}
    \caption{Subset ablation on VOC 1/4 segmentation (mIoU\,\%) under the same UniMatch/ResNet-101 $321\times321$ setting. A: CoVar-selected samples with MC$\ge0.95$; B: CoVar-selected samples with MC$<0.95$; C: CoVar-unselected samples with MC$\ge0.95$. Results are A=78.9, A+C (Fixed=0.95)=77.2, A+B (Ours)=79.8, and A+B+C=78.5.}
    \label{fig:partitioning-comparison}
\end{figure}

\textbf{Diagnostic Analysis of Pseudo-Label Selection.}
Across the matched ResNet-101 segmentation settings and the SimPLE-based classification setting, the selection diagnostics show that the adaptive confidence--variance weighting of CoVar translates the theoretical reliability conditions into practical selection gains.
When paired with the stronger DINOv2-B setting, CoVar continues to improve over the corresponding UniMatch~V2 baseline; the CLIP-B and MiT-B5 rows provide additional reference context rather than strict matched-backbone comparisons.
Fig.~\ref{fig:selrate} further shows the accuracy--coverage behavior during training.
In the early stage, the fixed threshold maintains high accuracy mainly by selecting fewer pseudo-labels, whereas CoVar expands coverage while quickly reaching comparable or higher accuracy.
In the middle and late stages, the fixed-threshold baseline gains coverage only gradually and its accuracy decreases, revealing the usual tradeoff between selecting more samples and preserving pseudo-label quality.
CoVar maintains higher coverage with stable accuracy, indicating that MC-RCV selection relaxes this tradeoff by admitting additional reliable predictions instead of simply lowering the threshold.

\begin{table}[t]
\centering
\caption{Comparison of prediction-partition strategies under matched settings: VOC reports mIoU (\%), while CIFAR-10 N100 and STL-10 N25 report Top-1 error rate (\%). t94/t95/t96 denote fixed thresholds with $\tau=0.94/0.95/0.96$.}
\label{tab:clustering_comparison}
\setlength{\tabcolsep}{1.6mm}{
\begin{tabular}{lcccc}
\toprule
Method & VOC 1/8 & VOC 1/4 & {N100} & {N25} \\
\midrule
t94          & 76.70 & 77.47 & 10.47 & 10.51 \\
t95          & 77.03 & 77.97 &  8.75 &  9.78 \\
t96          & 76.84 & 77.52 &  9.86 & 10.44 \\
k-means      & 77.13 & 78.06 &  7.54 & 10.29 \\
k-means+     & 77.92 & 78.73 &  6.12 &  9.93 \\
DBSCAN       & 78.21 & 78.99 &  5.63 &  9.28 \\
SVD (Ours)   & 78.94 & 79.81 &  4.75 &  8.74 \\
\bottomrule
\end{tabular}}
\end{table}

\subsection{Ablation Studies}

\textbf{Why Residual-Class Variance?}
We adopt MC and RCV because the CE decomposition indicates that reliable predictions should have both high confidence and low residual-class dispersion.
As shown in Table~\ref{tab:metric-combos}, the MC+RCV pair outperforms variants that add entropy, residual entropy, or margin features, suggesting that these additional handcrafted cues do not improve the current reliability embedding.
Fig.~\ref{fig:mc-rcv-evolution} provides a visual diagnosis: early predictions with high MC are still widely spread along RCV, while selected samples gradually move toward the high-MC/low-RCV region and concentrate near $(p_j(k'), v_j)=(1,0)$ by late training.
This explains the limitation of a fixed MC threshold, which can accept high-MC but high-RCV predictions and discard lower-MC but cleaner ones.
Fig.~\ref{fig:partitioning-comparison} confirms this subset effect: adding B improves mIoU from 78.9 to 79.8, whereas adding C degrades performance, showing that RCV helps recover useful MC$<0.95$ samples and reject unreliable MC$\ge0.95$ samples.
\begin{table}[t]
  \centering
  \caption{Comparison of different combinations of metrics. Ablation conducted on VOC augmented with SBD (321$\times$321) and CIFAR-10.}
  \setlength{\tabcolsep}{1mm}{
  \begin{tabular}{ccccc|cc|cc}
    \toprule
    $p_j(k')$  & $v_j$      & $H(p_j)$  &  $H_{res}$ & $m_j$       & 1/8    & 1/4              & N4    & N10         \\
    \midrule
    \checkmark & \checkmark &           &            &            & \textbf{78.9}&\textbf{79.8} & \textbf{4.81}&\textbf{4.71}\\
    \checkmark & \checkmark &\checkmark &            &            & 77.3 & 78.5                 & 8.26 & 7.08                \\
    \checkmark & \checkmark &           &\checkmark  &            & \underline{77.5} & \underline{79.1}                 & \underline{6.88} & \underline{6.42}                \\
    \checkmark & \checkmark &           &            & \checkmark & 76.9 & 78.2                 & 10.20 & 8.81                \\
    \checkmark & \checkmark & \checkmark& \checkmark & \checkmark & 71.5 & 76.3                 & 12.04 & 10.67                \\
    \checkmark &            & \checkmark&            &            & 70.7 & 72.4                 & 15.15 & 11.09                \\
    \checkmark &            &           & \checkmark &            & 71.6 & 74.6                 & 13.38 & 12.12                \\ 
    \checkmark &            &           &            & \checkmark & 69.4 & 71.9                 & 16.08 & 12.14                \\
    \bottomrule
    \end{tabular}
  }
  \label{tab:metric-combos}
\end{table}

\textbf{Comparative Analysis of Different Prediction Partitioning Strategies.}  
To evaluate robustness across prediction partitioning strategies, we compare SVD-based spectral relaxation with thresholding and clustering baselines on VOC augmented with SBD and representative classification benchmarks.
Results under consistent hyperparameters are summarized in Table~\ref{tab:clustering_comparison}. For the clustering baselines, the same two-dimensional reliability embedding $[\log p_j(k'),-g_j(p_j(k'))v_j]$ and the same surrounding training pipeline are retained; only the partitioning module is changed.
Across settings, SVD-based spectral relaxation consistently outperforms alternative partitioning strategies and shows stronger robustness to noise and complex feature distributions, thereby separating reliable pseudo-labels more effectively and providing stable gains in label-scarce scenarios.

\begin{table}[t]
  \centering
  \caption{Ablation study of different $\varepsilon$ values. Ablation conducted on VOC augmented with SBD (321$\times$321) and CIFAR-10.}
  \setlength{\tabcolsep}{2mm}{
  \begin{tabular}{c|cc|cc}
    \toprule
    $\varepsilon$ values    & 1/8    & 1/4    & N4  & N10   \\
    \midrule
    0.1                     & {77.9}& {78.2} & {7.92}& {7.05}\\
    0.01                    & \underline{78.1}& \underline{78.6} & \underline{7.18}& \underline{6.25}\\
    0.001                   & {77.5}& {78.0} & {8.45}& {7.72}\\
    0.0001                  & {77.4}& {77.8} & {8.92}& {7.18}\\
    0.00001                 & {77.3}& {77.6} & {9.35}& {8.55}\\
    0.000001                & {77.0}& {77.5} & {9.68}& {7.85}\\
    0.0000001               & {76.9}& {77.5} & {10.12}& {9.22}\\
    0.00000001              & {76.9}& {77.6} & {10.48}& {8.65}\\
    $(1 - p_i(k'))/(K - 1)$ & \textbf{78.9} & \textbf{79.8} & \textbf{4.81} & \textbf{4.71} \\
    \bottomrule
    \end{tabular}
  }
  \label{tab:epsilon-ablation}
\end{table}

\textbf{Comparative Analysis Under Different $\varepsilon$ Values.}  
The $\varepsilon$ ablation validates the adaptive choice $\varepsilon = (1 - p_i(k'))/(K - 1)$.
We compare this setting with fixed $\varepsilon$ values ranging from 1e-8 to 0.1, namely 1e-8, 1e-7, 1e-6, 1e-5, 1e-4, 1e-3, 0.01, and 0.1, on VOC augmented with SBD and CIFAR-10 under low-annotation ratios (1/8 and 1/4 for VOC; N4 and N10 for CIFAR-10), while maintaining consistent hyperparameters.
Results in Table~\ref{tab:epsilon-ablation} indicate that the adaptive $\varepsilon$ achieves the best performance across all splits. Very small fixed $\varepsilon$ values (1e-8 to 1e-5) lead to degradation due to numerical sensitivity, intermediate values (1e-4 and 1e-3) remain inferior to the adaptive choice, and larger $\varepsilon$ values (0.01 and 0.1) cause more pronounced drops because they deviate from the near-zero assumption. This adaptive setting balances theoretical fidelity and numerical stability, confirming its effectiveness for pseudo-label optimization.

\begin{table}[t]
  \centering
  \caption{Comparison of different terms in Eq.~\ref{eq14} for pseudo-label selection. Ablation conducted on VOC augmented with SBD (321$\times$321) and CIFAR-10.}
  \setlength{\tabcolsep}{1.9mm}{
  \begin{tabular}{ccc|cc|cc}
    \toprule
    $\overline{C}$    & $s\overline{RCV}$  & $Cov(g,v)$ & 1/8   & 1/4  & N4  & N10 \\
    \midrule
    \checkmark &           &            & \underline{77.9}   &\underline{78.5}   &\underline{6.85}     &\underline{5.72} \\
    \checkmark &\checkmark &            & 76.5   &77.7   &9.35     &8.05  \\
    \checkmark &           & \checkmark & 76.7   &78.3   &7.82     &6.55  \\
    \checkmark & \checkmark& \checkmark & \textbf{78.9} &\textbf{79.8} & \textbf{4.81} &\textbf{4.71} \\
               & \checkmark&            & 71.4   &73.1   &13.95     &9.80  \\
               & \checkmark& \checkmark & 74.6   & 76.6  & 11.65   & 9.40  \\ 
               &           & \checkmark & 73.4   & 75.8  & 15.25   & 10.35  \\
    \bottomrule
    \end{tabular}
  }
  \label{tab:batch-metric-combos}
\end{table}

\begin{table}[t]
  \centering
  \caption{Ablation study of nonlinear weighting coefficient $g_j(p_j(k'))$ designs. Ablation conducted on VOC augmented with SBD (321$\times$321) and CIFAR-10.}
  \setlength{\tabcolsep}{1.5mm}{
  \begin{tabular}{c|cc|cc}
    \toprule
    $g_j({{p}_{j}}({k}'))$ values                                     & 1/8   & 1/4    & N4  & N10   \\
    \midrule
    Entropy-Based: $c / H(p_j)$                                        & \underline{78.0}& {78.7} & \underline{5.50}& \underline{4.89}\\
    Clipped: $\min(w_o,w_c)$                                            & {77.6}& {77.9} & {8.10}& {6.72}\\
    NoWeight: $0$                                                      & {77.2}& {77.5} & {8.55}& {6.97}\\
    Linear: $1$                                                        & {77.8}& {78.1} & {9.53}& {7.24}\\
    Const: $\lambda{{{(K-1)}^{2}}}$                                         & {78.1}& {78.5} & {9.35}& {7.18}\\
    Alt-Rational: $\frac{{{(K-1)}^{2}}}{2(1-{{p}_{j}}({k}')+\delta)}$  & {77.9}& \underline{78.7} & {9.05}& {6.28}\\
    Full: $\frac{{{(K-1)}^{2}}}{2(1-{{p}_{j}}({k}'))}$ & \textbf{78.9} & \textbf{79.8} & \textbf{4.81} & \textbf{4.71} \\
    \bottomrule
    \end{tabular}
  }
  \label{tab:weighting-ablation}
\end{table}
\begin{figure*}[t]
  \centering
    \includegraphics[width=\linewidth]{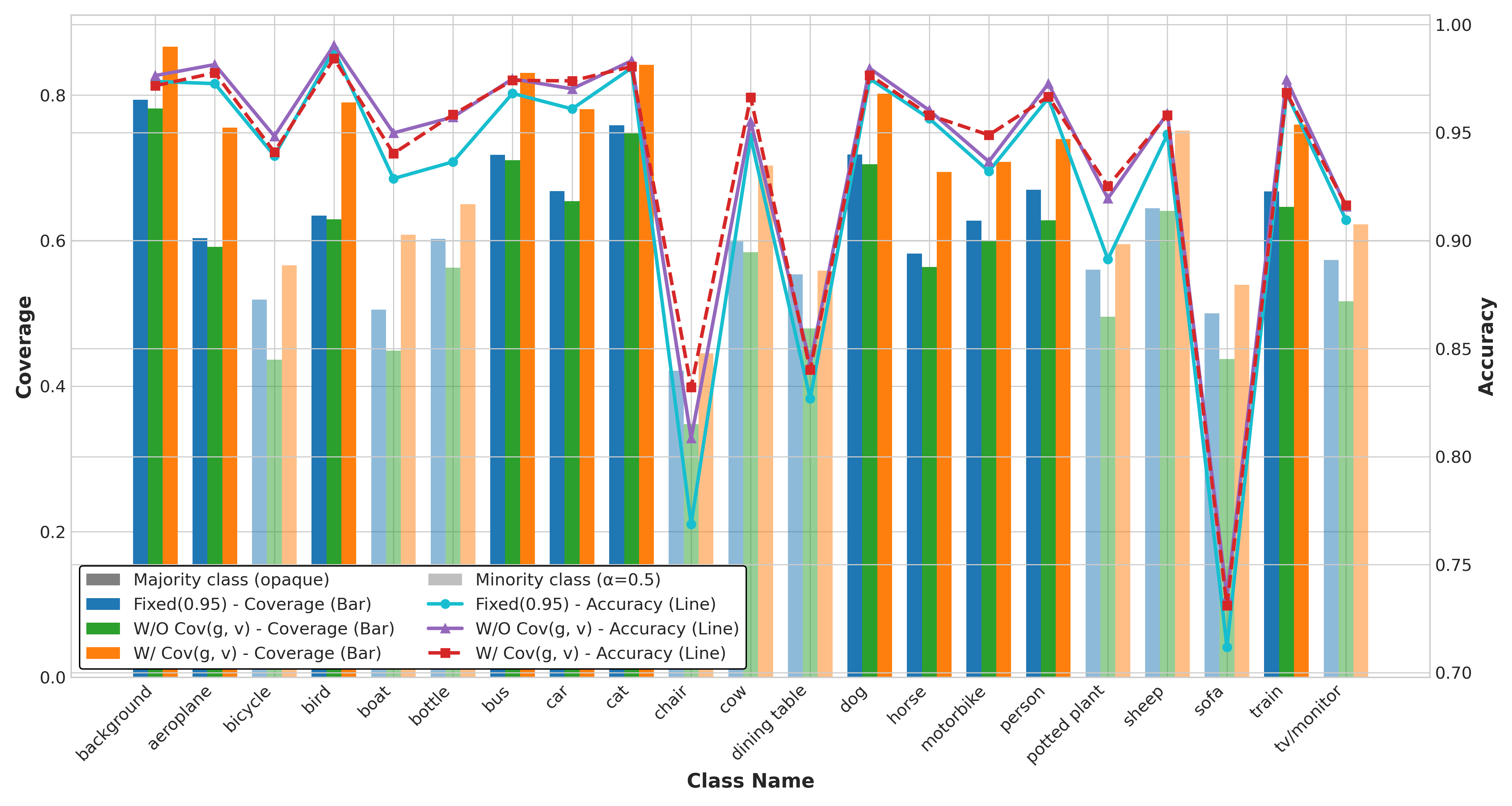}
    \caption{Class-wise pseudo-label coverage during training on VOC augmented with SBD (1/4, 321$\times$321).
    The fixed-threshold baseline over-selects majority classes and exhibits volatile selection for minority categories.
    Guided by the batch-level MC-RCV coupling summarized by $\mathrm{Cov}(g, v)$, CoVar yields smoother trajectories and more balanced coverage across classes by coupling high MC with low RCV.}
    \label{fig:class-selection-balance}
\end{figure*}

\textbf{Ablation Study of Nonlinear Weighting Coefficient $g_j(p_j(k'))$.}  
Table~\ref{tab:weighting-ablation} compares the proposed nonlinear weighting coefficient \(g_j(p_j(k'))=\frac{(K-1)^2}{2(1-p_j(k'))}\) with fixed, linear, clipped, entropy-based, and damped alternatives.
The Full design performs best across VOC and CIFAR-10, improving over the strongest alternative by +0.8/+1.1 mIoU on VOC and reducing CIFAR-10 error by 0.69/0.18 points at N4/N10.
Removing the weight or replacing it with fixed/linear forms degrades performance, showing that RCV must be penalized adaptively rather than treated as an unscaled feature.
Clipped, damped, and entropy-based variants are also weaker, indicating that the confidence-dependent penalty derived from the CE approximation better suppresses near-certain predictions with strong residual-class competition.

\textbf{Robustness Across Majority and Minority Classes.} 
Table~\ref{tab:batch-metric-combos} further examines robustness across majority and minority classes.
Using all three terms in Eq.~\ref{eq14} gives the best results, supporting the need for batch-level coupling between confidence and residual dispersion.
This coupling also explains Fig.~\ref{fig:selrate} and Fig.~\ref{fig:class-selection-balance}: compared with the fixed-threshold baseline, CoVar maintains higher coverage without sacrificing accuracy and yields smoother, more balanced head--tail coverage.
By suppressing overconfident high-RCV samples and admitting reliable minority-class samples that confidence-only rules under-select, CoVar mitigates long-tail pseudo-label bias without explicit class-frequency reweighting.

\begin{figure}[t]
  \centering
    \includegraphics[width=\linewidth]{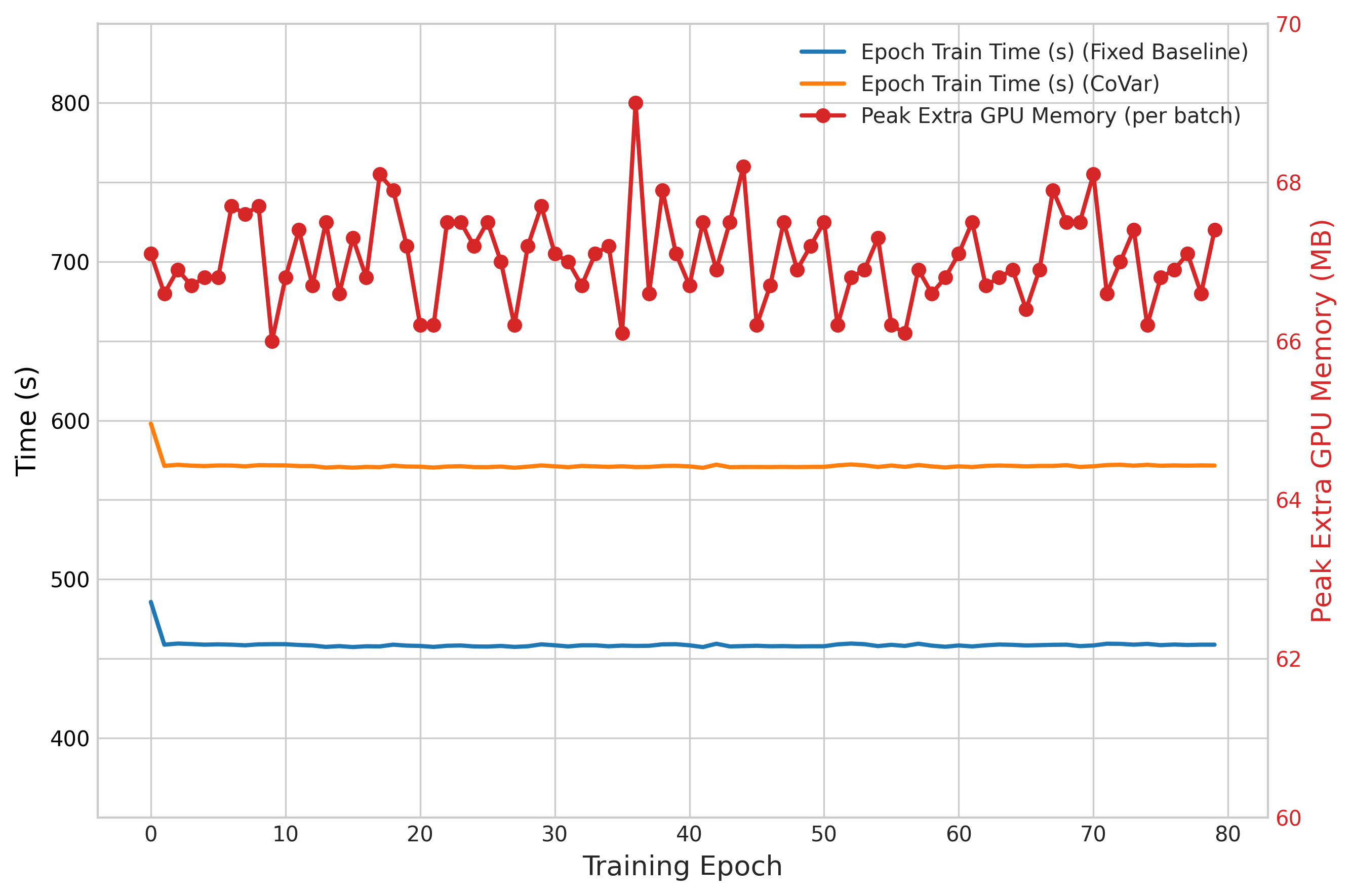}
    \caption{Per-epoch training time and peak extra GPU memory introduced by CoVar on VOC augmented with SBD under the 1/4 labeled split (ResNet-101 backbone).}
    \label{fig:overhead}
\end{figure}

\textbf{Training Overhead.}  
Fig.~\ref{fig:overhead} reports the per-epoch training time and peak extra GPU memory introduced by CoVar on VOC augmented with SBD under the 1/4 labeled split (ResNet-101 backbone).
Without CoVar, the baseline requires approximately 460 s per epoch; with CoVar enabled, the per-epoch cost rises to roughly 570 s, corresponding to an overhead of approximately 24\%.
This 24\% increase reflects the full training-time CoVar pipeline rather than the isolated SVD partition alone, including embedding construction, Gaussian weighting, mask generation, synchronization, and the associated loss computation.
The additional peak GPU memory per batch remains consistently below 70 MB throughout all 80 training epochs, with negligible variance across epochs.
These results confirm that CoVar imposes a moderate and predictable training overhead (\(\approx\)24\%) for SVD-based reliability estimation.

\begin{figure}[t]
  \centering
    \includegraphics[width=\linewidth]{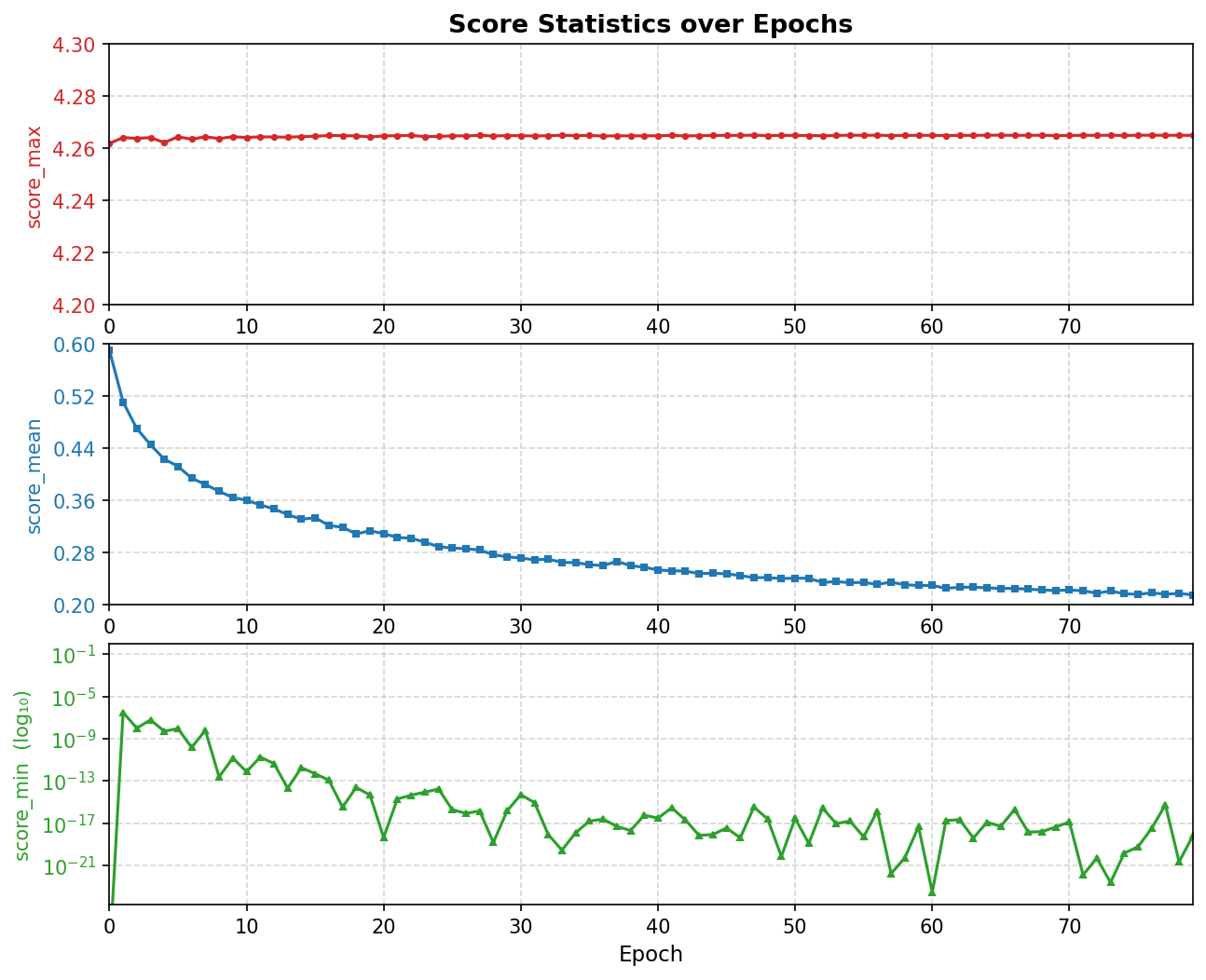}
    \caption{Evolution of the theoretically derived score $\text{Score} = -\log p_j(k') +  g_j \cdot v_j$ across training epochs.}
    \label{fig:score-evolution}
\end{figure}

\textbf{Score Evolution.}  
To motivate the adoption of SVD-based spectral relaxation over a fixed score threshold, Fig.~\ref{fig:score-evolution} visualizes the evolution of the theoretically derived score $\text{Score} = -\log p_j(k') +  g_j \cdot v_j$ across training epochs.
The score maximum remains nearly constant at $\approx 4.26$ throughout all epochs, while the score mean decreases monotonically from $\approx 0.60$ at epoch 0 to $\approx 0.21$ by epoch 75, reflecting the progressive improvement of pseudo-label confidence as the model converges.
Most critically, the score minimum (displayed on a $\log_{10}$ scale) fluctuates erratically over more than 14 orders of magnitude, ranging from $\sim\!10^{-21}$ to $\sim\!10^{-7}$, without any stable trend.
This extreme and unpredictable dynamic range, particularly in the lower tail of the distribution, demonstrates that no fixed, universal threshold can reliably separate reliable pseudo-labels from noisy ones across different training stages.
Any static cutoff would either prematurely discard informative samples in early epochs or retain corrupted ones in later stages.
This observation empirically validates the need for an adaptive selection strategy. SVD-based spectral relaxation addresses this issue by detecting structural discontinuities on the data manifold, thereby identifying the natural ``fault line'' between reliable and unreliable pseudo-labels without presupposing a fixed score boundary.

\subsection{Limitations}

CoVar estimates pseudo-label reliability from batch-level MC-RCV statistics, and its behavior can therefore depend on the composition of each mini-batch, especially when reliable samples from rare classes are scarce.
Although the method improves head-tail pseudo-label coverage in our experiments, it is not an explicit class-rebalancing strategy and does not impose class-frequency constraints.
CoVar also operates on current predictions only and does not maintain temporal memory for samples that may become reliable later in training.
Finally, our validation focuses on semi-supervised semantic segmentation and image classification; extending the same formulation to detection, domain adaptation, or other SSL settings may require task-specific adaptations.

\section{Conclusion}

This paper presented CoVar, a confidence--variance framework that moves pseudo-label selection beyond confidence-only thresholding.
From entropy minimization, we derived a second-order CE decomposition showing that reliable pseudo-labels require both high MC and low RCV, with a confidence-dependent variance penalty and a batch-level interpretation of long-tail bias mitigation.
CoVar turns this criterion into a practical training-time module by separating predictions in a confidence--variance space with SVD-based spectral relaxation and converting the separation into sample weights.
Across semi-supervised segmentation and classification benchmarks, CoVar consistently improves strong baselines while adding no inference-time overhead.
Future work will extend this reliability criterion to broader SSL settings, including domain adaptation and other uncertainty-aware tasks.

%%===========================================================================================%%
%% Declarations                                                                             %%
%%===========================================================================================%%

\section*{Statements and Declarations}

\textbf{Acknowledgements.}
This work was supported in part by the National Natural Science Foundation of China (No.~62503505), the China Postdoctoral Science Foundation (No.~GZC20251179, 2025T012HN), and the Hunan Natural Science Foundation (No.~2025JJ60424).

\textbf{Competing interests.}
The authors declare that they have no competing interests.

\textbf{Data availability.}
All datasets used in this study are publicly available.

\textbf{Code availability.}
The source code for CoVar is publicly available at \url{https://github.com/ljs11528/CoVar_Pseudo_Label_Selection}.

\textbf{Author contributions.}
J.L. conceived the method, implemented the framework, conducted experiments, and wrote the manuscript.
P.L. and L.H. supervised the project and revised the manuscript.
All authors reviewed and approved the final manuscript.

\textbf{Supplementary information.}
The online version contains supplementary material.
Supplementary material provides theoretical proofs (derivations of Eqs.~\ref{eq10}--\ref{eq13} and stability/monotonicity analysis related to Eq.~\ref{eq15}), a low-complexity implementation with pseudocode for the CoVar module, and additional experimental results and visualizations.

\bibliography{ref}% common bib file

\end{document}

\title[CoVar Supplementary Material]{Supplementary Material for CoVar: Confidence--Variance-Guided Pseudo-Label Selection for Semi-Supervised Learning}

% \author[1]{\fnm{Jinshi} \sur{Liu}}

% \author*[2]{\fnm{Pan} \sur{Liu}}

% \author*[3]{\fnm{Lei} \sur{He}}

% \email{brianpanliu@hkust-gz.edu.cn}
% \email{helei\_xb@hnust.edu.cn}

% \affil[1]{\orgdiv{College of Artificial Intelligence}, \orgname{Shenzhen University}, \orgaddress{\city{Shenzhen}, \postcode{518060}, \country{China}}}

% \affil*[2]{\orgdiv{Information Hub}, \orgname{Hong Kong University of Science and Technology (Guangzhou)}, \orgaddress{\city{Guangzhou}, \postcode{510006}, \country{China}}}

% \affil*[3]{\orgdiv{School of Information and Electrical Engineering}, \orgname{Hunan University of Science and Technology}, \orgaddress{\city{Xiangtan}, \postcode{411100}, \country{China}}}

%%Journal name for SI header
%% International Journal of Computer Vision (IJCV)



\maketitle

\section{Overview}
This supplementary material provides derivations for Eqs.~\ref*{main-eq10}--\ref*{main-eq13}, stability and monotonicity analyses related to Eq.~\ref*{main-eq15}, the binary-class special case, a normalized-partition justification of the spectral objective used for reliability separation (Sec.~\ref{sec:proof}), a low-complexity implementation of CoVar with pseudocode (Sec.~\ref{sec:algorithm}), and extended implementation details and experimental comparisons (Sec.~\ref{sec:add_experiment}).

\section{Discussion of the CE Decomposition and Spectral Partition Objective}
\label{sec:proof}
\subsection{Justification of the Second-Order Approximation} 

Most existing SSL methods select pseudo-labels primarily by the maximum confidence $p_j(k')$, which corresponds to a zeroth- or first-order characterization of the cross-entropy. 
Specifically, the cross-entropy for a prediction $p_j$ can be written as
\begin{equation}
\mathrm{CE}(q_j,p_j) = -\log p_j(k') - \sum_{k\ne k'} q_j(k)\log p_j(k).
\end{equation}

Under the assumption that residual-class probabilities are approximately uniform, the second term is treated as a constant, leading to the widely used approximation
\begin{equation}
\mathrm{CE}(q_j,p_j) \approx -\log p_j(k').
\end{equation}

This approximation ignores the internal distribution of residual classes and therefore fails to distinguish predictions with identical confidence but different uncertainty structures.

To capture finer-grained reliability, we perform a second-order Taylor expansion of $\log p_j(k)$ around the uniform residual distribution. 
This yields the approximation
\begin{equation}
\mathrm{CE}(q_j,p_j) \approx -\log p_j(k') + \frac{(K-1)^2}{2(1-p_j(k'))}\,v_j,
\end{equation}
where $v_j$ denotes the variance of residual-class probabilities. 
Compared with confidence-only selection, this formulation introduces an additional degree of freedom, enabling discrimination between predictions that share the same maximum confidence but differ in class-wise dispersion.

The second-order approximation is sufficient for our reliability analysis.
Extending the expansion to higher orders introduces terms of the form $\sum_{k\ne k'} \delta_j^3(k)$, which depend on higher-order moments of the residual distribution. 
Such quantities are difficult to estimate reliably in practice and lack clear interpretability in the context of prediction reliability.
They also introduce numerical instability, especially when $p_j(k') \to 1$.

Therefore, the second-order approximation provides a practical balance:
it is expressive enough to capture the key uncertainty structure through $v_j$, while remaining stable and computationally tractable.

This analysis shows that fixed-threshold methods implicitly rely on a low-order approximation of cross-entropy, whereas CoVar uses a higher-order characterization. 
Thus, confidence-based thresholding is inherently limited by the low-order nature of its approximation.
By incorporating residual-class variance, CoVar provides a more informative criterion for pseudo-label selection without introducing unnecessary complexity.

\subsection{Stability Justification of the Taylor Approximation}
\label{sup:taylor_stability}

This subsection addresses a possible concern about applying Taylor expansion when
residual-class probabilities become very small. We do not require a uniformly exact
expansion for all predictions. Instead, we only need the approximation to be stable
in the regime where CoVar trusts pseudo-labels: high confidence with small and not
overly uneven residual probabilities.

For sample \(j\), let \(\alpha_j=p_j(k')\), \(r_j=1-\alpha_j\),
\(m=K-1\), and \(\mu_j=r_j/m\). For each residual class \(k\neq k'\), write
\begin{equation}
\delta_j(k) = p_j(k)-\mu_j,
\qquad
\sum_{k\neq k'} \delta_j(k)=0.
\end{equation}
The expansion is taken around the positive adaptive center \(\mu_j\), rather than
around zero:
\begin{equation}
\begin{aligned}
\log p_j(k)
&=
\log \mu_j
+
\frac{\delta_j(k)}{\mu_j}
-
\frac{\delta_j^2(k)}{2\mu_j^2}
+
R_{j,k}, \\
R_{j,k}
&=
\frac{\delta_j^3(k)}{3\xi_{j,k}^3},
\quad
\xi_{j,k}\in I_{j,k},\\
I_{j,k}
&=
\bigl(\min\{p_j(k),\mu_j\},\,
\max\{p_j(k),\mu_j\}\bigr).
\end{aligned}
\end{equation}
Thus, as long as \(\alpha_j<1\), we have \(\mu_j>0\) and the expansion is not
performed at the singular point of \(\log x\). In implementation, probabilities are
also clipped by a small constant \(\rho>0\) to avoid exact zeros.

The quantity that enters the CE approximation is the residual-weighted remainder,
not the unweighted Taylor remainder alone. Under the adaptive choice
\(\epsilon=\mu_j\), assume the residual probabilities are locally balanced, i.e.,
\(|\delta_j(k)|\leq \eta\mu_j\) for all \(k\neq k'\) and some \(\eta\in(0,1)\).
Then \(\xi_{j,k}\geq(1-\eta)\mu_j\), and
\begin{equation}
\epsilon \sum_{k\neq k'} |R_{j,k}|
\leq
\frac{m\eta^3}{3(1-\eta)^3}\mu_j
=
\mathcal{O}(\mu_j).
\end{equation}
Therefore, as \(\alpha_j\rightarrow 1\) (equivalently \(\mu_j\rightarrow 0\)), the
weighted Taylor remainder converges to zero.

This argument is intentionally local. If the residual mass is highly uneven, the
local balance condition may fail, but then the residual-class variance
\(v_j=\frac{1}{m}\sum_{k\neq k'}\delta_j^2(k)\) becomes large and CoVar treats the
sample as less reliable. Hence the second-order approximation is stable precisely in
the regime where the method is intended to trust pseudo-labels.

\subsection{\texorpdfstring{$f(\cdot)$ Monotonicity under the adaptive choice of \(\epsilon\)}{f(.) Monotonicity under the adaptive choice of epsilon}}
The monotonicity claim in the main paper is used only under the adaptive setting
\(\epsilon = \mu_j = \frac{1-p_j(k')}{K-1}\). Let \(\alpha = p_j(k')\). Since
\(\alpha = \max_k p_j(k)\) for a valid \(K\)-class probability vector, we always
have \(\alpha \geq \frac{1}{K}\).

Substituting \(\epsilon = \frac{1-\alpha}{K-1}\) into
\(f_j(\alpha) = \log \alpha - (K-1)\epsilon \log \frac{(K-1)\alpha}{1-\alpha}\)
gives
\begin{equation}
f(\alpha) = \log \alpha - (1-\alpha)\log \frac{(K-1)\alpha}{1-\alpha}.
\end{equation}
Differentiating yields
\begin{equation}
f'(\alpha) = \log \frac{(K-1)\alpha}{1-\alpha}.
\end{equation}
For \(\alpha \in (\frac{1}{K},1)\), we have
\(\frac{(K-1)\alpha}{1-\alpha} > 1\), and hence \(f'(\alpha) > 0\); at
\(\alpha = \frac{1}{K}\), we have \(f'(\alpha)=0\). Therefore, \(f(\alpha)\) is
strictly increasing on \([\frac{1}{K},1)\).

Under the same adaptive choice,
\begin{equation}
g(\alpha) = \frac{(K-1)^2}{2(1-\alpha)},
\end{equation}
which is also strictly increasing on \([\frac{1}{K},1)\).

\subsection{Normalized Partition View of the Spectral Objective}
For the reliability-partition problem in semi-supervised semantic segmentation, we divide predictions into two clusters: a high-reliability cluster and a low-reliability cluster. Let \(S\in\{0,1\}^{N\times2}\) be the corresponding two-way reliability assignment matrix, with \(\sum_c S_{n,c}=1\) and \(c\in\{1,2\}\) indexing reliability clusters rather than semantic classes. Each prediction embedding is \(\tilde{h}_n\in\mathbb{R}^{2}\), and these embeddings are stacked row-wise as
\begin{equation}
X=[\tilde{h}_1^T,\tilde{h}_2^T,\ldots,\tilde{h}_N^T]^T\in\mathbb{R}^{N\times2}.
\end{equation}
Equivalently, under the column-wise notation used in the main paper, \(\Phi=[\tilde{h}_1,\tilde{h}_2,\ldots,\tilde{h}_N]\in\mathbb{R}^{2\times N}\) and \(X=\Phi^T\).

The derivation starts from the standard normalized kernel k-means / normalized partition objective obtained by minimizing the within-cluster reconstruction error
\begin{equation}
\mathcal{L}(S)=\|X-S(S^TS)^{-1}S^TX\|_F^2.
\end{equation}
Let \(D=S^TS=\operatorname{diag}(n_1,n_2)\) and define the normalized indicator matrix
\begin{equation}
Y=SD^{-1/2},
\qquad Y^TY=I.
\end{equation}
Then
\begin{equation}
S(S^TS)^{-1}S^T=YY^T,
\end{equation}
so the objective becomes
\begin{equation}
\mathcal{L}(S)=\|X-YY^TX\|_F^2.
\end{equation}
Using the Frobenius norm identity and \((YY^T)^2=YY^T\), we obtain
\begin{equation}
\mathcal{L}(S)=\|X\|_F^2-\operatorname{Tr}(Y^TXX^TY).
\end{equation}
Since \(X=\Phi^T\), this is equivalent to maximizing
\begin{equation}
J_{\mathrm{norm}}(S)=\operatorname{Tr}\left(D^{-1/2}S^T\Phi^T\Phi SD^{-1/2}\right).
\end{equation}
This is the standard normalized kernel k-means / normalized association criterion induced by the kernel \(K=\Phi^T\Phi\). Importantly, the normalization by \(D^{-1/2}\) cannot be dropped, because the cluster sizes \(n_1\) and \(n_2\) depend on \(S\) and are therefore not constants.

The discrete optimization over \(S\) remains NP-hard. Relaxing \(Y\) from a normalized indicator matrix to an arbitrary orthonormal matrix \(U\in\mathbb{R}^{N\times2}\) with \(U^TU=I\) yields the spectral relaxation
\begin{equation}
\max_{U^TU=I}\operatorname{Tr}(U^T\Phi^T\Phi U),
\end{equation}
whose solution is given by the first two eigenvectors of \(\Phi^T\Phi\) by the Ky Fan theorem. Therefore, the SVD-based spectral relaxation used in the main paper should be interpreted as a spectral heuristic motivated by this normalized partition objective. The simplified spectral form in the main text is a compact presentation of the relaxed criterion, not a claim of strict equivalence to the unnormalized objective \(\operatorname{Tr}(S^T\Phi^T\Phi S)\).

\subsection{Binary-Class Special Case}
\label{sup:binary_case}

When \(K=2\), the residual set contains only one class. Let \(\alpha=p_j(k')\) denote the maximum confidence. Then the only residual probability is \(1-\alpha\), and its residual mean is
\begin{equation}
\mu_j=\frac{1-\alpha}{K-1}=1-\alpha.
\end{equation}
Hence the residual deviation satisfies
\begin{equation}
\delta_j(k)=p_j(k)-\mu_j=0,\qquad k\ne k',
\end{equation}
and the residual-class variance becomes
\begin{equation}
v_j=\frac{1}{K-1}\sum_{k\ne k'}\delta_j^2(k)=0.
\end{equation}
Therefore, the variance term in Eq.~\ref*{main-eq12} vanishes:
\begin{equation}
\mathrm{CE}(q_j,p_j)\approx -f_j(\alpha).
\end{equation}
Under the adaptive choice \(\epsilon=\mu_j\), \(f_j(\alpha)\) is increasing on \([\frac{1}{2},1)\), as shown above. Thus, in binary classification, minimizing the approximate CE is equivalent to maximizing MC. The RCV component becomes informative only in the multi-class setting \(K\geq 3\), where the residual mass can be distributed unevenly across multiple competing classes.

\section{Algorithm}
\label{sec:algorithm}

\subsection{Low-complexity Implementation of Confidence--Variance Spectral Separation (CVSS)}

In the spectral partition step of the main paper, we require the right singular vectors of $\Phi \in \mathbb{R}^{2 \times N}$
(where $N = HW$ for segmentation), which coincide with the eigenvectors of
$\Phi^T\Phi \in \mathbb{R}^{N \times N}$.
A naive approach that explicitly forms $\Phi^T\Phi$ and then performs eigendecomposition costs
$\mathcal{O}(N^2)$ in memory and $\mathcal{O}(N^3)$ in time, making it infeasible at the pixel level.
We instead use the fact that $\Phi$ has only two rows, which reduces the overall cost to $\mathcal{O}(N)$.

\paragraph{Key identity.}
Because $\Phi \in \mathbb{R}^{2 \times N}$ has rank at most 2, its economy (thin) SVD takes the form
\begin{equation}
\Phi = \mathbf{U}\,\boldsymbol{\Sigma}\,\mathbf{V}^T,
\mathbf{U} \in \mathbb{R}^{2\times 2},
\boldsymbol{\Sigma} \in \mathbb{R}^{2\times 2},
\mathbf{V} \in \mathbb{R}^{N\times 2}.
\end{equation}
Substituting into $\Phi^T\Phi$ and using orthogonality of $\mathbf{U}$:
\begin{equation}
\begin{aligned}
\Phi^T\Phi
&= \mathbf{V}\,\boldsymbol{\Sigma}^T\mathbf{U}^T\mathbf{U}\boldsymbol{\Sigma}\,\mathbf{V}^T \\
&= \mathbf{V}\,\boldsymbol{\Lambda}\, \mathbf{V}^T, \\
\boldsymbol{\Lambda}
&= \operatorname{diag}(\sigma_1^2,\sigma_2^2).
\end{aligned}
\end{equation}
Thus, the columns of $\mathbf{V}\in\mathbb{R}^{N\times 2}$ are exactly the two non-trivial
eigenvectors of $\Phi^T\Phi$, and they are all we need for the spectral assignment.

\paragraph{$\mathcal{O}(N)$ implementation via $\Phi\Phi^T$.}
Rather than computing the economy SVD of $\Phi$ directly (which still requires an
$\mathcal{O}(N)$ pass over all columns), we use the small left factor $\Phi\Phi^T \in \mathbb{R}^{2\times 2}$.
The implementation first forms $\mathbf{M} = \Phi\Phi^T \in \mathbb{R}^{2\times 2}$ by accumulating $N$ rank-1 outer products,
$\mathbf{M} = \sum_{n=1}^{N} \mathbf{h}_n \mathbf{h}_n^T$.
This requires $\mathcal{O}(4N) = \mathcal{O}(N)$ multiplications and $\mathcal{O}(N)$ memory for $\Phi$.
It then eigendecomposes the $2\times 2$ matrix $\mathbf{M} = \mathbf{U}\boldsymbol{\Lambda}\mathbf{U}^T$, which has a closed-form solution and costs $\mathcal{O}(1)$, independent of $N$.
Finally, the right singular vectors are recovered as $\mathbf{V} = \Phi^T \mathbf{U}\boldsymbol{\Sigma}^{-1}$ through two matrix-vector products of size $N$, with cost $\mathcal{O}(4N) = \mathcal{O}(N)$.
Total time complexity: $\mathcal{O}(N)$; total memory: $\mathcal{O}(N)$ (storing $\Phi$ and $\mathbf{V}$,
both of size $2N$).
This contrasts with the naive $\Phi^T\Phi$ route, which requires $\mathcal{O}(N^2)$ memory and $\mathcal{O}(N^3)$ time.
In practice, for a typical 513$\times$513 segmentation map ($N \approx 2.6\times10^5$), this linear-time subroutine completes in milliseconds, whereas the $\mathcal{O}(N^3)$ route is intractable.
This complexity analysis only covers the CVSS spectral partition itself; the full training-time overhead reported in the main paper also includes feature construction, Gaussian weighting, mask generation, synchronization, and loss computation, while inference remains unchanged because CVSS is used only during training.

\subsection{Pseudocode for Confidence--Variance Spectral Separation} 

Algorithm~\ref{alg:1} provides the pseudocode for Confidence--Variance Spectral Separation (CVSS).
For implementation convenience, Algorithm~\ref{alg:1} uses zero-based cluster indices \(\{0,1\}\), which are equivalent to the \(\{1,2\}\) cluster notation used in the main text.
Using a spectral separation strategy based on the joint confidence--variance embedding, CoVar excludes many high-confidence false predictions during pseudo-label selection and improves performance in semi-supervised semantic segmentation.

\definecolor{commentgreen}{RGB}{0, 128, 0} % 
\algrenewcommand\algorithmiccomment[1]{\textcolor{commentgreen}{\footnotesize\texttt{\# #1}}}
\begin{algorithm}
\caption{Simplified pseudocode for Confidence--Variance Spectral Separation (CVSS).}
\label{alg:1}
\begin{algorithmic}[1]
\Function{CVSS}{$p_{max}$, $v_j$, $K$}
  \State \Comment{Build the confidence--variance embedding}
  \State $h_1 = \log(p_{max})$
  \State $h_2 = -\dfrac{(K-1)^2}{2\,(1-p_{max})} \, v_j$
  \State $\Phi = \operatorname{stack}([h_1,\, h_2],\;\text{dim}=0)$
  \State \Comment{Spectral partition}
  \State $U,\,\Sigma,\,V^T = \operatorname{svd}(\Phi,\;\texttt{full\_matrices=False})$
  \State $V = (V^T)^T$
  \State $S = \operatorname{argmax}(\operatorname{abs}(V),\;\text{dim}=1)$
  \State \Comment{Cluster statistics}
    \State ${stats} = []$
    \For{$c$ in $\{0, 1\}$}
        \State $\Phi_c = \Phi[:,\; S == c]$
    \State $\mu_c = \Phi_c.\operatorname{mean}(\text{dim}=1)$
    \State $\sigma_c = \Phi_c.\operatorname{std}(\text{dim}=1)$
        \State ${stats}.\operatorname{append}((\mu_c,\,\sigma_c))$
    \EndFor
  \State \Comment{Select the reliable cluster}
    \State ${scores} = []$
    \For{$(\mu_c, \sigma_c)$ in ${stats}$}
    \State ${scores}.\operatorname{append}(\mu_c[0] + \mu_c[1])$
    \EndFor
    \State $c^{*} = \operatorname{argmax}({scores})$
    \State $(\mu,\,\sigma) = {stats}[c^{*}]$
  \State \Comment{Compute sample weights}
    \State ${weight} = \exp\!\left(-\dfrac{(\Phi - \mu[:,\,\text{None}])^{2}}{2\,\sigma[:,\,\text{None}]^{2}}\right)$
  \State ${weight} = {weight}.\operatorname{prod}(\text{dim}=0)$
    \State ${weight}[(S == c^{*}) \;\&\; (\Phi[0,:] > \mu[0]) \;\&\; (\Phi[1,:] > \mu[1])] = 1$
    \State \Return ${weight}$
\EndFunction
\end{algorithmic}
\end{algorithm}
\begin{figure*}[t]
  \centering
   \includegraphics[width=1\linewidth]{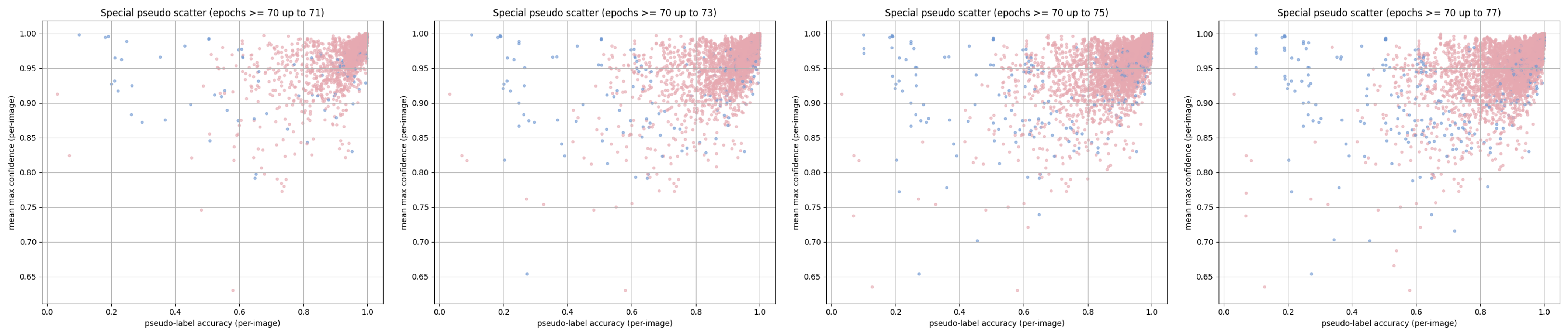}
   \caption{Evolution of pseudo-label selection during training on PASCAL VOC 2012 (1/8, 513$\times$513). Each scatter plot shows maximum confidence versus average pseudo-label accuracy at different epochs. Majority-class samples (warmer colors) initially dominate high-confidence regions, while minority-class samples (cooler colors) progressively converge to the high-accuracy quadrant as the joint MC-RCV criterion of CoVar refines selection, achieving balanced coverage by epoch 77.} 
  \label{fig:sup_pseudo_scatter}
\end{figure*}
\section{Extensive Experiment Details}
\label{sec:add_experiment}  

\subsection{More Implementation Details}

For semantic segmentation, we follow the UniMatch setting used in the main paper: weak augmentations include random scaling in $[0.5,2.0]$, cropping, and flipping, while strong augmentations add ColorJitter, random grayscale, Gaussian blur, and CutMix.
We also use weak views and a modified strong view without random grayscale or Gaussian blur.
Following the CorrMatch-style setting, we apply 50\% random channel dropout as feature perturbation and use BF16 mixed-precision training.

For image classification, we follow SimPLE with a Wide ResNet-28-2 backbone. CIFAR-10 and CIFAR-100 use class-balanced fixed-seed splits for labeled, validation, and unlabeled subsets. Weak augmentation consists of random crop with reflective padding of $1/8$ image size, random horizontal flip, and normalization; strong augmentation further applies RandAugmentNS$(n=2,m=10)$, while validation and test use normalization only. We set $K_{\text{strong}}=7$ for CIFAR-10 and $4$ for CIFAR-100. Following the released SimPLE CIFAR setup, we use SGD with Nesterov momentum $0.9$, learning rate $0.03$ with per-iteration cosine decay, weight decay $5\times10^{-4}$ for CIFAR-10 and $10^{-3}$ for CIFAR-100, and EMA decay $0.999$. The objective is $\mathcal{L}=\mathcal{L}_x+\lambda_u\mathcal{L}_u+\lambda_{\mathrm{pair}}\mathcal{L}_{\mathrm{pair}}$, with $\lambda_u=\lambda_{\mathrm{pair}}=75$ on CIFAR-10 and $150$ on CIFAR-100, similarity threshold $0.9$, and fixed confidence threshold $\tau_c=0.95$ inherited from SimPLE rather than tuned by CoVar. CoVar is inserted after pseudo-label guessing and before $\mathcal{L}_u$ and $\mathcal{L}_{\mathrm{pair}}$; here \(N=B_u\times K_{\text{strong}}\) denotes the augmentation views entering the loss.

\subsection{More Experimental Results}

Fig.~\ref{fig:sup_pseudo_scatter} visualizes the evolution of pseudo-label selection during training on PASCAL VOC 2012 (1/8, 513$\times$513), plotting maximum confidence (MC) versus average accuracy from epoch 70 to epoch 77.
Early in this interval, majority-class samples (warmer colors) dominate the high-MC regions, while minority-class samples (cooler colors) remain dispersed.
As training proceeds, minority-class samples move toward the high-MC/high-accuracy quadrant because the joint MC-RCV criterion identifies reliable tail-class samples.
By the final epochs, both groups converge with lower dispersion.
This behavior shows that $\mathrm{Cov}(g,v)$ suppresses overconfident yet unreliable predictions, enabling smoother and more equitable selection than fixed-threshold baselines.

\subsection{More Visualizations}

\begin{figure*}[t]
  \centering
   \includegraphics[width=1\linewidth]{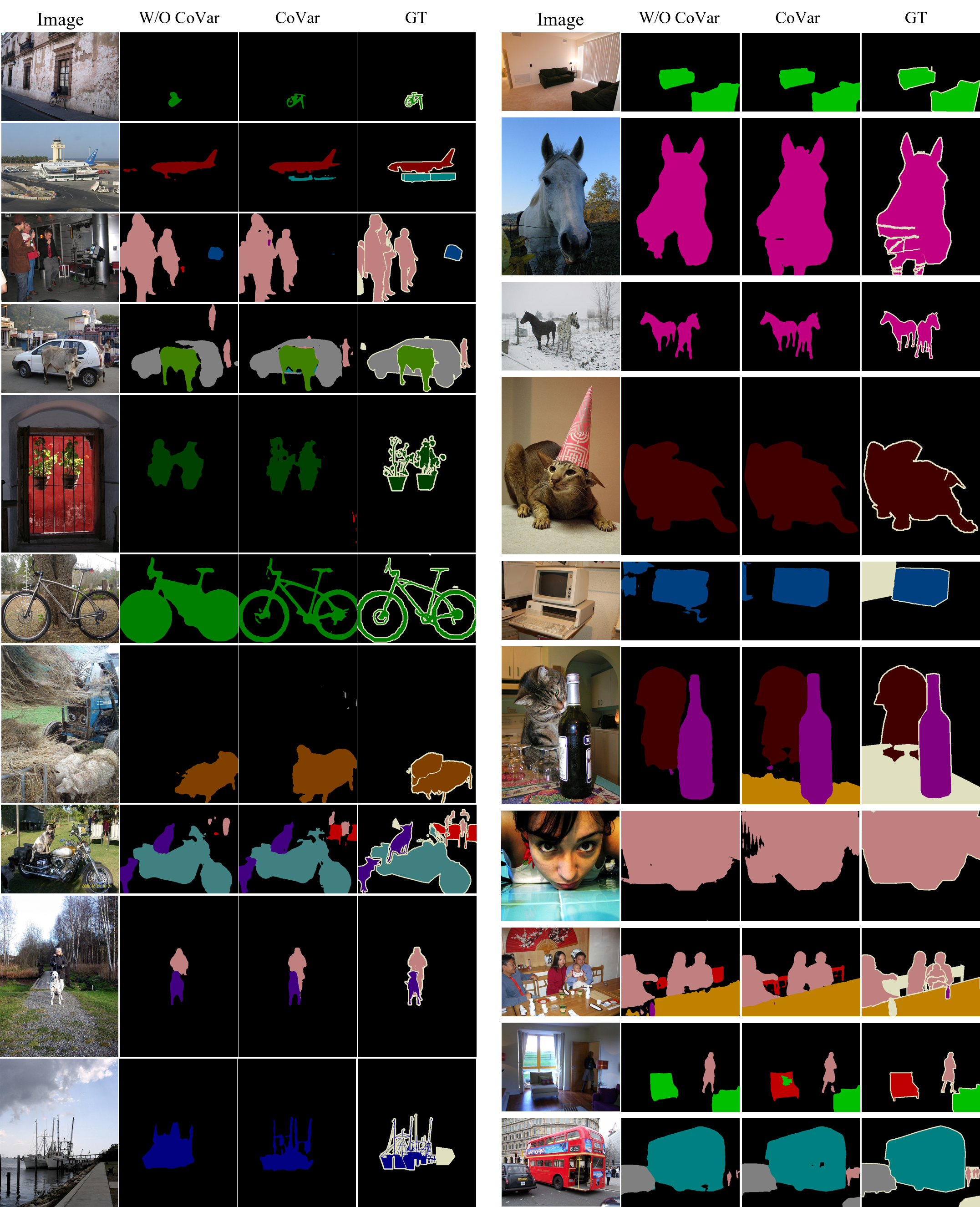}
   \caption{Qualitative comparison of segmentation results on PASCAL VOC 2012 (1/8, 513$\times$513). Top rows show input images and ground truth; middle rows show the UniMatch V1 baseline without CoVar; bottom rows show the proposed method with CoVar. CoVar yields sharper boundaries, reduced false positives, and improved completeness for minority classes compared with the baseline.} 
  \label{fig:sup_qualitative}
\end{figure*}

Fig.~\ref{fig:sup_qualitative} compares qualitative segmentation results on PASCAL VOC 2012 (1/8, 513$\times$513) between the UniMatch V1 baseline without CoVar and the proposed method with CoVar.
CoVar produces sharper object boundaries, especially around fine structures such as limbs and wheels, where the baseline often gives blurred or fragmented edges.
It also suppresses spurious regions predicted by the baseline and improves completeness for tail classes such as bicycles and potted plants.
These gains corroborate the quantitative mIoU improvements by enforcing joint MC-RCV reliability.

